\documentclass[
reprint,
amsmath,amssymb,
aps,
pra,
]{revtex4-1}

\usepackage{graphicx}
\usepackage[caption=false]{subfig}
\graphicspath{{img/}}
\usepackage{float}
\usepackage{booktabs} 
\usepackage{xr}

\newcommand{\tabincell}[2]{\begin{tabular}{@{}#1@{}}#2\end{tabular}}
\begin{document}

\title{Triad State Space Construction for Chaotic Signal Classification with Deep Learning}%
\author{Yadong Zhang}
\author{Xin Chen}%
 \email{xin.chen.nj@xjtu.edu.cn}
\affiliation{%
 Center of Nanomaterials for Renewable Energy, \\ State Key Laboratory of Electrical Insulation and Power Equipment, \\School of Electrical Engineering, \\ Xi'an Jiaotong University, Xi'an 710054, Shaanxi, China 
}%

\date{\today}

\begin{abstract}
Inspired by the well-known permutation entropy (PE), an effective image encoding scheme for chaotic time series, Triad State Space Construction (TSSC), is proposed. The TSSC image can recognize higher-order temporal patterns and identify new forbidden regions in time series motifs beyond the Bandt-Pompe probabilities. The Convolutional Neural Network (ConvNet) is widely used in image classification. The ConvNet classifier based on TSSC images (TSSC-ConvNet) are highly accurate and very robust in the chaotic signal classification.
\end{abstract} 

\maketitle 

\textit{Introduction}---The time series analysis, particularly the time series classification and prediction\cite{timeclassification,forecasting_1990}, has broad applications in science and engineering such as medicine\cite{medicine_2015}, brain study\cite{EEG_2006}, transportation\cite{traffic_2014}, power systems\cite{power_energy_2012}, finance\cite{finance_1996} and {\it etc}.  There are various real-world datasets such as human-activity sensing data, speech signals, and electroencephalogram (EEG). Using the algorithms to reveal the temporal structures and clustering patterns will allow us to understand and interpret the real-world time series\cite{TSC_2001} and their underlying dynamics.

Practical time series classification methods based on similarity measurement or features representation, such as Euclidean Distance (ED), Dynamic Time Wrapping (DTW)\cite{DTW_1994}, Shapelet\cite{lines_shapelet_2012}, Bag of SFA Symbols\cite{schafer_2014} and {\it etc}, have been used extensively. Recently machine learning and pattern recognition algorithms are extended in the time series prediction\cite{TSP_2007} and classification\cite{TSC_2001}. The deep multilayer perceptrons(MLP), Fully-connected Convolutional Network (FCN) and residual networks (ResNet)\cite{wang_time_2016} are able to classify the time series in the UCR time series archive\cite{UCRArchive2018}. 
Recently machine learning\cite{flame_2019}\cite{3_body_2019} is applied for model-free prediction of spatiotemporally chaotic systems purely from observations of the system’s past evolution. 

Understanding the temporal structures in time series can improve the ability of forecasting, classification and anomaly detection\cite{timeclassification,timeseriespre}.
The time series motif clustering technique discovers temporal structures in the time series, such as motifs\cite{motifs_2003}, periodic patterns\cite{periodic_motif_1999}, partially ordered patterns\cite{partially_order_pattern_2006}. Bandt and Pompe\cite{bandt_permutation_2002} introduced the Permutation Entropy (PE) by assigning a probability distribution to the time series motifs. The chaotic dynamics characterization\cite{rosso_characterization_2013} can be analysed with PE. Based on the Bandt-Pompe probability, the complexity-entropy causality plane can distinguish deterministic and stochastic dynamics\cite{rosso_characterization_2013}. However, the classification is not very successful since the PE and triadic time series motifs only have the information about ordinal orders. 

In addition to ordinal orders, the temporal details of time series, such as amplitude, also contain important information. The time series classification approach with deep learning is developed recently by encoding the time series into images with the methods such as Recurrence Plots\cite{RP_1987,CNN_image_classification_2017}, delay coordinate reconstruction (DCR)\cite{DCR_1980,DCR_2010}, Gramian Angular Summation/Difference Fields (GASF/GADF) and Markov Transition Fields (MTF)\cite{GAF_MTF_2015}, {\it etc.}  However, we found that the deep learning classifier based on these image encoding methods are not able to accurately classify chaotic time series. 

In this letter, a novel image encoding method of time series is proposed. The construction of the triadic state space is the extension of the ordinal orders of triadic time series motifs. We refer to the image encoding method as the Triadic State Space Construction (TSSC). We can classify the chaotic time series with the ConvNet using the corresponding images generated by TSSC. The coarse-graining heat-map of the TSSC image is prepared as the input to ConvNet. Using the surrogate time series of nine chaotic maps, we test the performance of the ConvNet classifier. Since chaotic maps and its surrogate time series are characterized with three factors, control parameter, initial condition and segmentation, the training and test datasets are prepared accordingly. We test the performance of the ConvNet classifier using three experiments on the datasets. 

\textit{Triad State Space Construction}---For a time series $\{x_t, t=1,2,3,...,N\}$,  a sequence of triadic time series motifs (TTSM)\cite{xie_time_2019} is defined as $T_t:\{(x_t, x_{t+1}, x_{t+2}), t=1, 2, 3, ..., N-2\}$.  There are six TTSM ordinal orders, $\{231, 213, 123, 132, 213, 321\}$ in PE\cite{triad_3_2012}. For example, given a TTSM $T = (8,2,5)$, its ordinal pattern is $231$ since $x_2 < x_3 < x_1$.  Given the sequence of TTSM $T_t$, the polar coordinates, $R_t$ and $\theta_t$, of TTSM at $t$,  are defined as,
\begin{eqnarray}
R_t = \sqrt{\hat{x}_t^2 + \hat{y}_t^2}\\
\theta_t = arctan(\frac{\hat{y}_t}{\hat{x}_t})&, \theta_t \in [-\pi, \pi)\nonumber
\end{eqnarray}\label{equ:Rtheta}
where $\hat{x}_t=x_{t+1}-x_t$ and $\hat{y}_t=x_{t+2}-x_{t+1}$. 
Instead of using Cartesian coordinates, the polar coordinates are used due to the limited boundary of $[-\pi, \pi)$ for $\theta_t$. According to the polar coordinates, the TTSM sequence can be transformed into a pseudo triad state space where 
the distance can be calculated with the TTSM coordinates. 
Taking for example three TTSM instances, $T_1=(1.9, 2.0, 3.0), T_2=(2.1, 2.0, 3.0), T_3=(2.9, 2.0, 3.0)$ whose polar coordinates are $(1.005, 1.471), (1.005, 1.670), (1.345, 2.303)$. Apparently, $T_1$ and $T_2$ in the TSSC image are close to each other although $T_1$ belongs to the ordinal order of $123$ and $T_2$ and $T_3$ to $231$. The polar coordinate apparently provides another dimension of information beyond the ordinal order patterns. 
\begin{figure}[H]
\centering
 \subfloat[]{\includegraphics[width=0.080\textwidth]{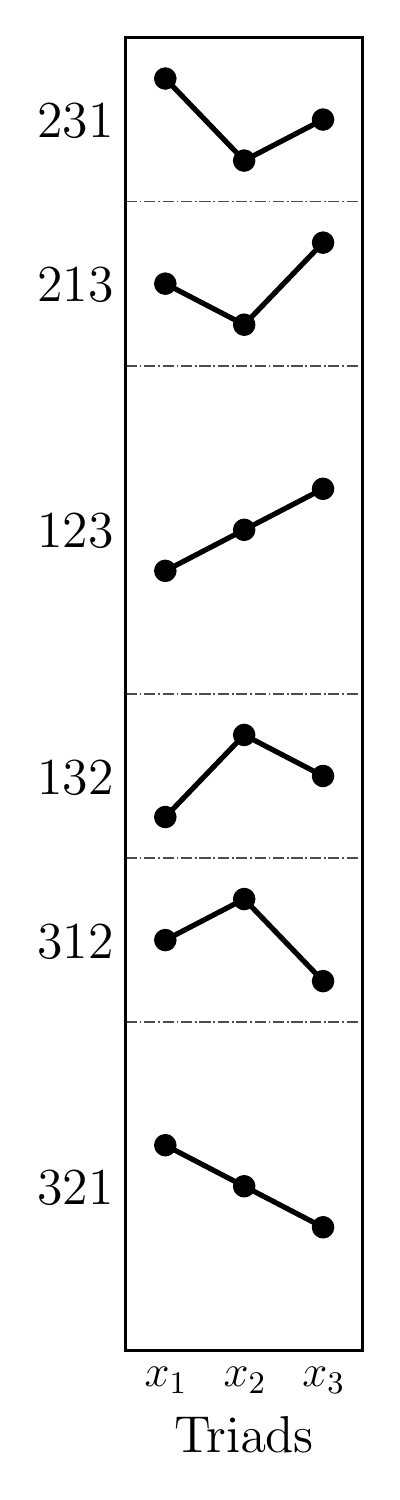}\label{psr:motif}}
 \subfloat[]{\includegraphics[width=0.075\textwidth]{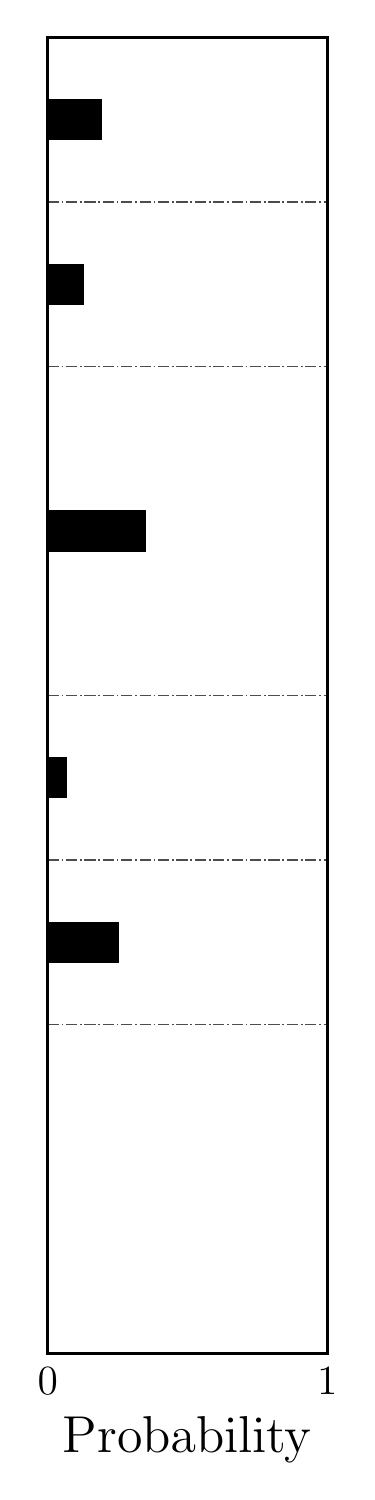}\label{psr:freq}}%
 \subfloat[]{\includegraphics[width=0.345\textwidth]{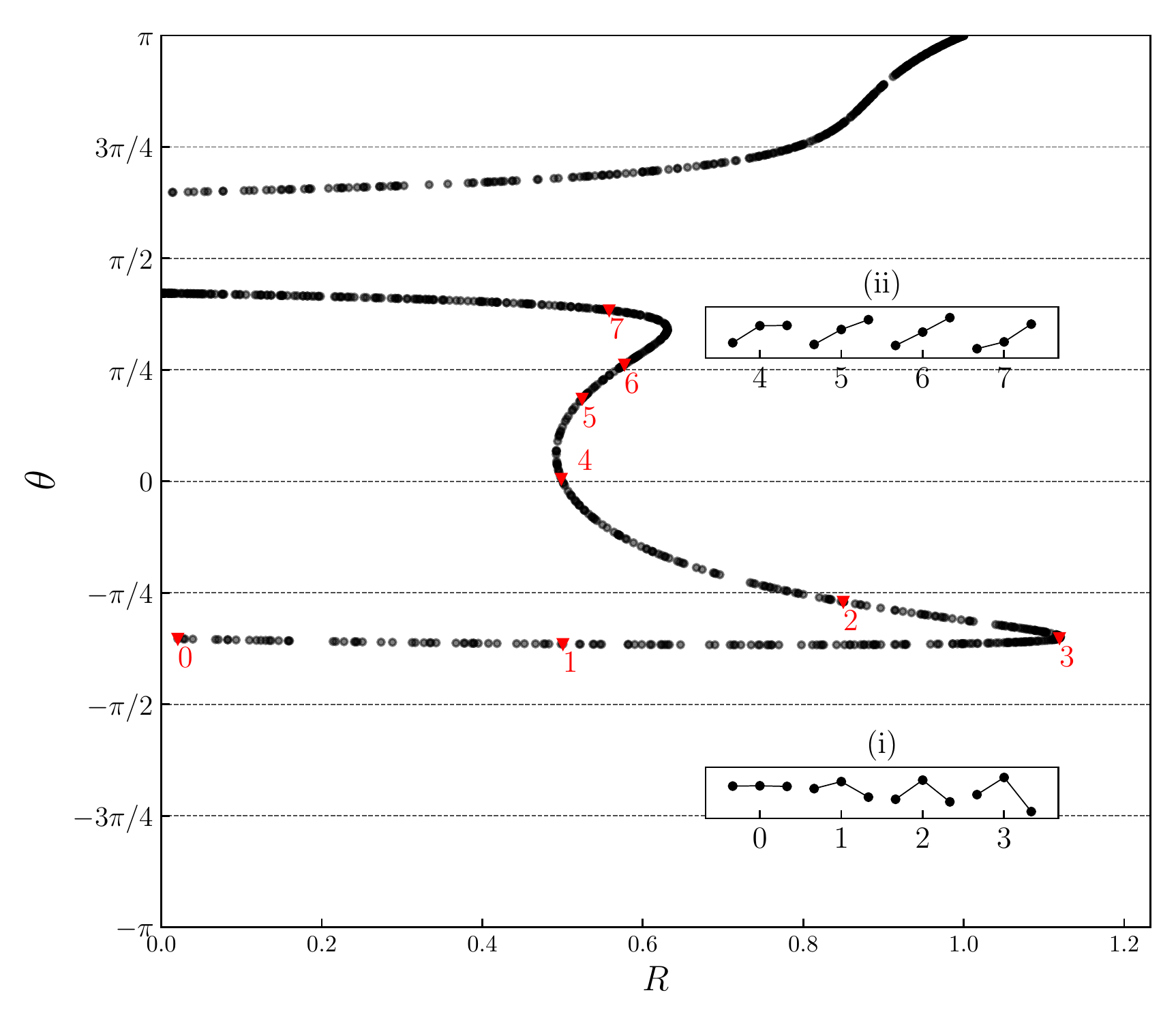}\label{psr}}%
\caption{
(a) Six TTSM ordinal orders. (b) Bandt-Pompe probabilities of the six TTSM ordinal orders. (c) The TSSC image in which Box (i) are four TTSM instances with the ordinal order of $312$; Box (ii) are four TTSM instances with the ordinal order of $123$. The surrogate time series are generated by the logistic map
$x_{t+1}=r x_t(1-x_t)$ that has the parameter $r=4.0$ and initial state $x_1=1e-6$. The length of the time series is 1000. 
}\label{fig:psr}
\end{figure} 
For a time series and its TTSM sequence, TSSC offers an image representation of Bandt-Pompe probabilities. 
Figure~\ref{fig:psr} shows the TSSC image of the surrogate time series of the chaotic Logistic map\cite{may_1976} in comparison with the Bandt-Pompe probabilities. 
Figure~\ref{psr:motif} shows the six TTSM instances of the six ordinal orders. Figure~\ref{psr:freq} shows the Bandt-Pompe probabilities of the six TTSM ordinal orders. Figure~\ref{psr} show the six regions corresponding to the six TTSM ordinal orders in the triad state space. 
The TSSC image shows two new forbidden regions around $\theta = \pi/2$ and $-\pi/2$, which are not reflected in the Bandt-Pompe probabilities. The TSSC image shows that eight TTSM instances in Box(i) and Box(ii).
Box (i) in Figure~\ref{psr} shows the four TTSM instances with the ordinal order of $312$ while Box (ii) shows the four TTSM instances with the ordinal order of $123$.  

\begin{figure}[H]
\centering
\subfloat[TSSC Image]{\includegraphics[width=0.23\textwidth]{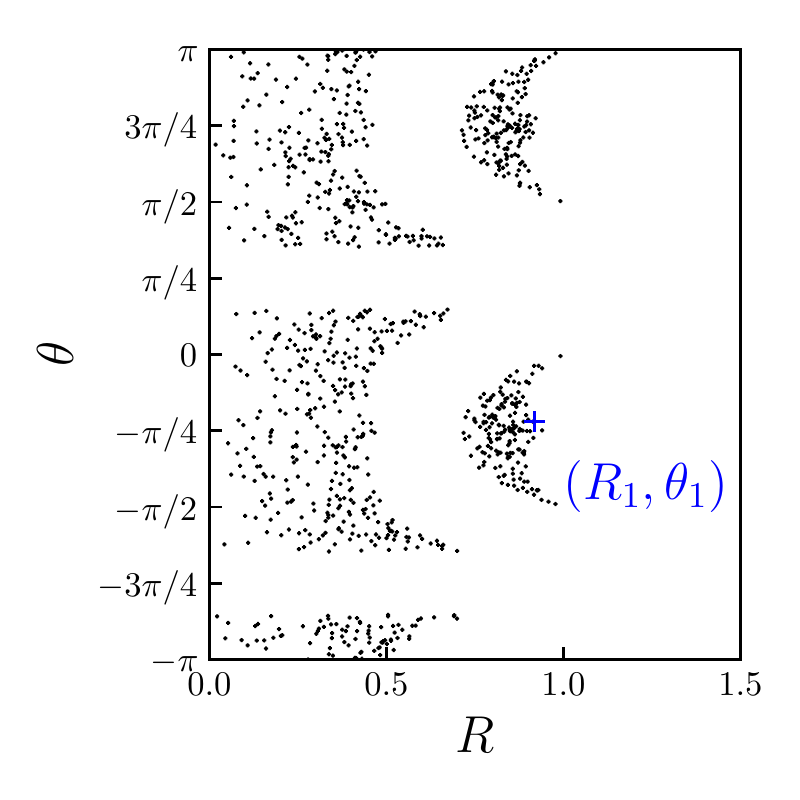}\label{tss-anode}}
\subfloat[DRC Image]{\includegraphics[width=0.23\textwidth]{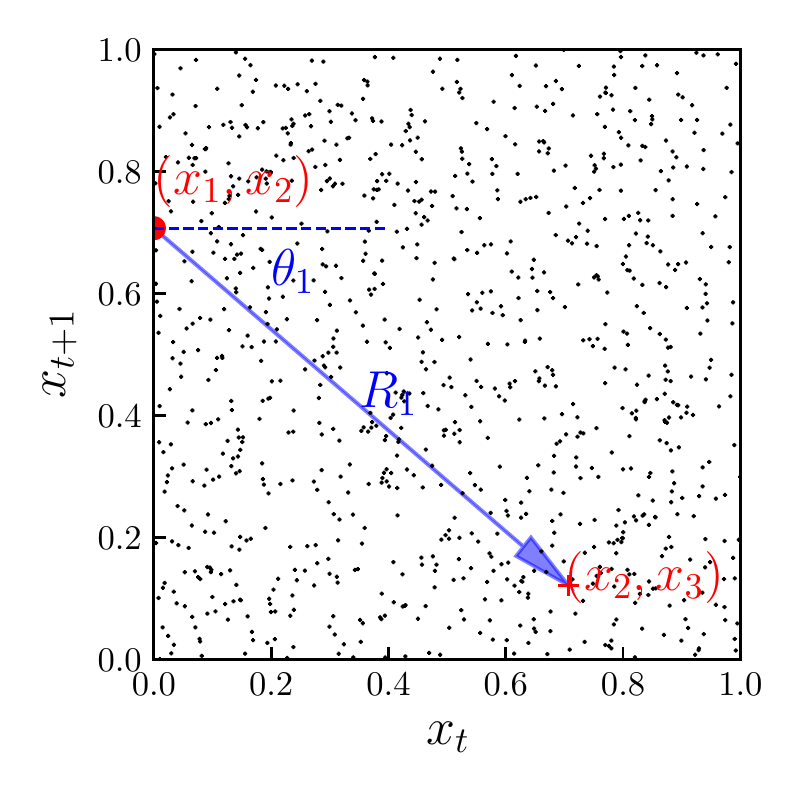}\label{dcr-anode}}
\caption{(a) The TSSC image of the surrogate time series $x_t$ of Annode's cat map. (b) The DCR image (time delay embedding is 1). The vector from $(x_1, x_2)$ to $(x_2, x_3)$ in the DCR image is equivalent to the point, $(R_1, \theta_1)$ in the TSSC image. The length of the time series is 2000.}\label{fig:compareDelay}
\end{figure}

\begin{figure}[h]
\centering
\subfloat[The coarse-graining grid]{\includegraphics[width=0.23\textwidth]{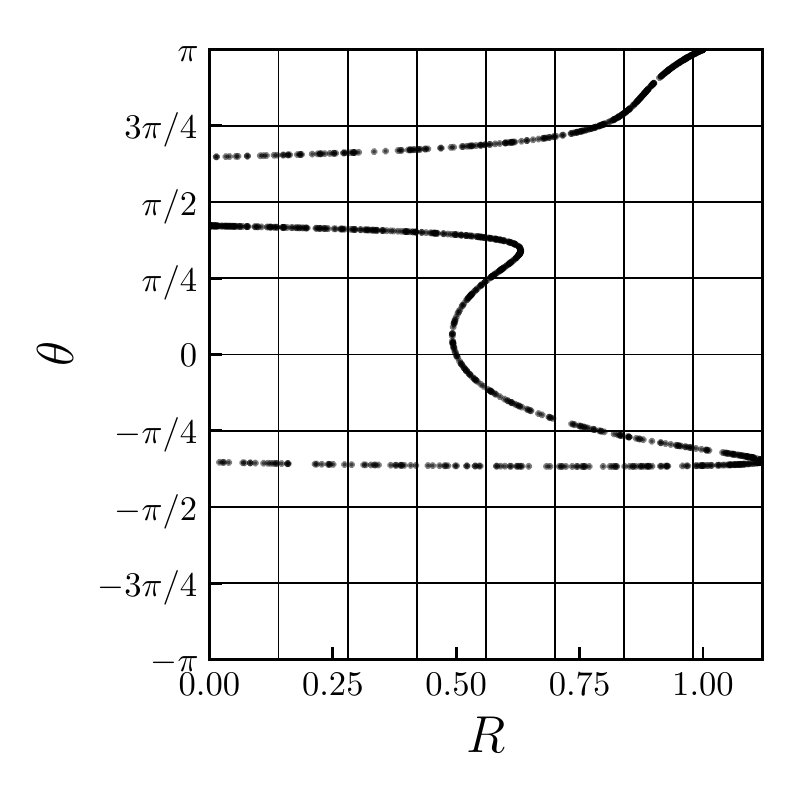}\label{greyscale}}
\subfloat[The heat-map]{\includegraphics[width=0.23\textwidth]{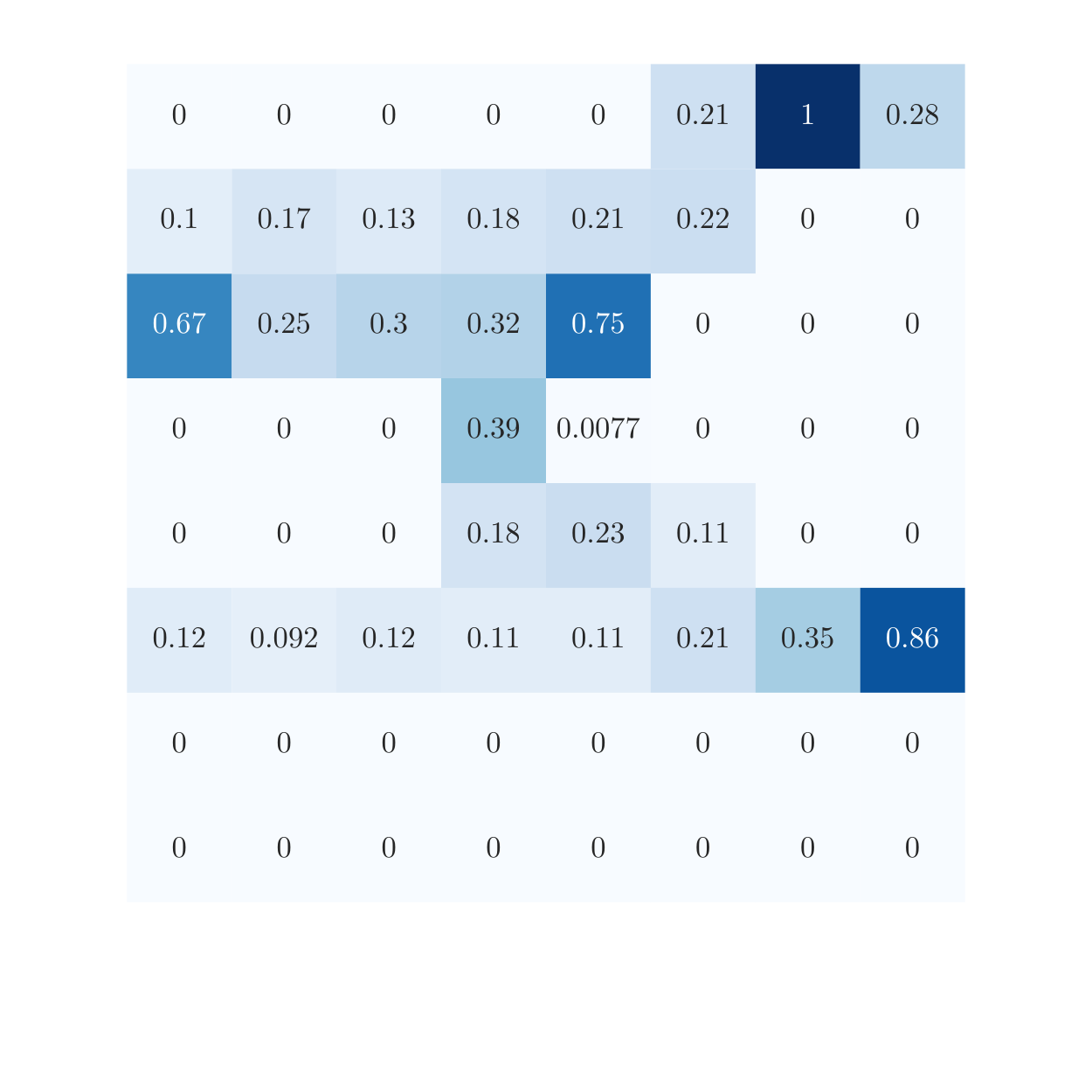}\label{heatmap}}
\caption{The coarse-graining heat-map of the TSSC image.}\label{fig:pixel}
\end{figure}

In Figure~\ref{fig:compareDelay}, the TSSC image is compared with the DCR image (time delay embedding is 1) using the surrogate time series of Annode's cat map, $x_{t+1}=x_t+y_t \pmod{1}$, $y_{t+1}=x_t+k y_t \pmod{1}$ which has the control parameter $k=2$ and the initial conditions, $x_1=0$, $y_1=1/\sqrt{2}$. The comparison show that the TSSC image has very unique structural pattern while the DCR one is totally blurry. A point in the TSSC image in Figure~\ref{tss-anode}, $(R_1, \theta_1)$ represents a triad, ($x_1,x_2, x_3$), that is equivalent to the vector involving the two points, ($x_1$, $x_2$) and ($x_2$, $x_3$) in the DCR image in Figure~\ref{dcr-anode}. The TSSC images in Figures~\ref{fig:chaotic-TSS} in Supplemental Material display distinct and unique patterns for the surrogate time series of nine chaotic maps. However the corresponding DCR images in Figures~\ref{fig:chaotic-DCR} in Supplemental Material don't display strong patterns. The TSSC image provides a new way to represent the triad clustering in time series. Potentially, due to the unique patterns in the TSSC images, the similarity measure can be defined accordingly\cite{image_similarity_1998}. 
\textit{Convolutional Neural Network Classifier with TSSC Images}---The Convolutional Neural Network (ConvNet)\cite{CNN_1998} is a class of deep neural networks, most commonly applied to analyze visual imagery. ConvNet demonstrates the capability of the pattern recognition\cite{CNN_pattern_1962} and image classification\cite{CNN_image_classification_2017}.  Inspired by the success of ConvNet in images classification and the triad state space patterns, we demonstrate that ConvNet is capable of classifying the time series with the TSSC images accurately. We call the new deep learning classifier, TSSC-ConvNet. As the input of ConvNet, the coarse-graining heat-map of the TSSC image is generated by counting how many triads and normalizing the number in the grid squares as shown in Figure~\ref{greyscale}. Figure~\ref{fig:pixel} shows the $8\times8$ heat-map of the TSSC image for the surrogate time series of the logistic map. 
The TSSC-ConvNet structure is shown in Figure~\ref{fig:heatmap-convnet} in Supplemental Material with the heat-map as the input. In this letter, the $64\times64$ heat-maps are used. To compare the TSSC images with the DCR images for the ConvNet classification, a similar heat-map of the DCR image is generated. The benchmark methods are DCR-ConvNet whose structure is the same as TSSC-ConvNet, and the TS-ConvNet structure is shown in Figure~\ref{fig:ts-convnet} in Supplemental Material.

\begin{figure}[H]
\centering
\includegraphics[width=0.50\textwidth]{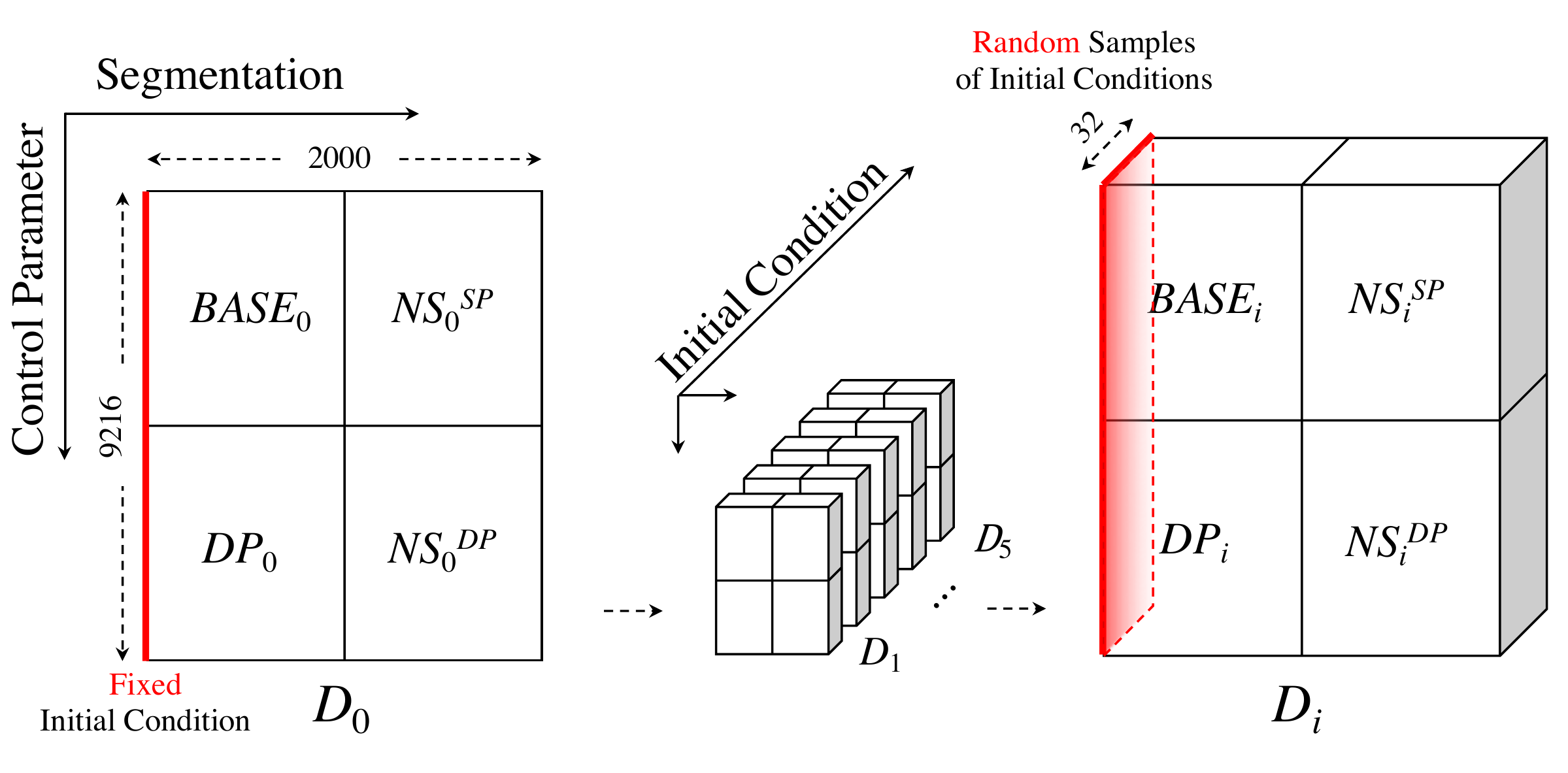}
\caption{The datasets $D_i$ have three dimensions of control parameter, initial condition, and segmentation. In $D_0$, there are 9216 time series generated with the nine chaotic maps. Each chaotic map has 1024 control parameters and one fixed initial condition. In $D_i,\; i=1,2,3,4,5$, each chaotic map has 1024 control parameters and 32 random samples of initial conditions. All the datasets $D_i\; i=0,1,2,3,4,5$, are segmented into four quadrants, $BASE_i$, $DP_i$, $NS^{SP}_i$ and $NS^{DP}_i$ for the training and test of ConvNet.}\label{fig:dataset}
\end{figure}

\textit{Datasets and Experiments}---To understand and benchmark the performance of TSSC-ConvNet classifier, nine dynamic chaotic maps\cite{chaos_analysis_2003} are used to generate surrogate time series for the training and test datasets. The nine chaotic maps are listed in Table~\ref{tab:TSsetting} in Supplemental Material. In the datasets, the surrogate time series are labeled with the nine chaotic maps. The percentage of the time series in the test dataset correctly assigned to the chaotic maps is used to quantify the classification accuracy. The chaotic maps have control parameters and initial conditions. The surrogate time series in the dataset are generated with different control parameters and initial conditions. In addition, the segmentations of the chaotic dynamic time series are considered since it can make the classification complicated.  the training and test datasets have the three dimensions of control parameter, initial condition and segmentation as shown in Figure~\ref{fig:dataset}. 

Figure~\ref{fig:dataset} shows the six datasets, $D_i$, $i=0,1,2,3,4,5$.  The $D_0$ dataset only has one slice that has 9216 times series with the length of 2000 generated by the nine chaotic dynamic maps. All the datasets $D_i,\;i=1,2,3,4,5$ consist of 32 slices. The initial conditions and control parameters of the nine chaotic maps in $D_0$ are listed in Table~\ref{tab:TSsetting} in Supplemental Material, where for each chaotic map, one fixed initial condition is chosen and 1024 control parameters are sampled uniformly in their allowable ranges. In total, there are 9216 control parameters corresponding to the 9216 time series in the sole slice of $D_0$.
Different from $D_0$, the remaining five datasets of $D_i,\, i=1,2,3,4,5$ have 32 slides with the initial conditions randomly sampled from the uniform distributions $[c_0, c_0+p_i)$ where $p_i=0.1,0.2,0.3,0.4,0.5$ and $c_0$ is the initial conditions chosen in the sole slice of $D_0$.
All the six datasets $D_i$ are segmented into the training and test datasets to train and test ConvNet for the three classifier, TS-ConvNet, DCR-ConvNet and TSSC-ConvNet. Figure~\ref{fig:dataset} shows the four segmented quadrants, $BASE_i$, $DP_i$, $NS^{SP}_i$ and $NS^{DP}_i$ in $D_i$.
$BASE_i$ is the base dataset for the training and $DP_i$ is the dataset for the test in the first experiment on the control parameters.
Segmentation is important to test three classifiers.  $NS^{SP}_i$ is the test dataset for the test in the third experiment on the segmentations with the same control parameters as $BASE_i$. $NS^{DP}_i$ is for the test in the third experiment on the segmentation with different control parameters from $BASE_i$. 
In total, each slice in the datasets, $BASE_i$, $DP_i$, $NS^{SP}_i$ and $NS^{DP}_i$ has 4608 time series with the length of 1000. With the datasets, we will carry out three experiments for the effects of control parameters, initial conditions and segmentations of time series to evaluate and benchmark the accuracies of the three classifier, TS-ConvNet, DCR-ConvNet and TSSC-ConvNet.

\textit{Results for Control Parameters}---In the first experiment to understand the effect of control parameter, the datasets $BASE_i$ are used for the training of ConvNet and $DP_i$ for the test and benchmark of the classification accuracy. The time series in $BASE_i$ and $DP_i$ all have the same initial conditions.
With the trained ConvNet based on the six datasets, $BASE_i$ respectively, the classification accuracies of the three classifiers, TS-ConvNet, DCR-ConvNet and TSSC-ConvNet are evaluated. Table~\ref{tab:DP} in Supplemental Material shows all the three classifiers all have the classification accuracies above 90{\%} for all the six test datasets $DP_i$. TSSC-ConvNet has the highest accuracies above 99{\%}. With ConvNet trained with $BASE_i$, the classification accuracies according to $DP_i$ are very similar to the three classifiers. The first experiment tells that if the time series in the training and test datasets only have different control parameters for each chaotic maps, ConvNet can have very high classification performance using the inputs of the raw time series, DCR images, and TSSC images.

\begin{figure}[H]
\centering
\subfloat[]{\includegraphics[width=0.23\textwidth]{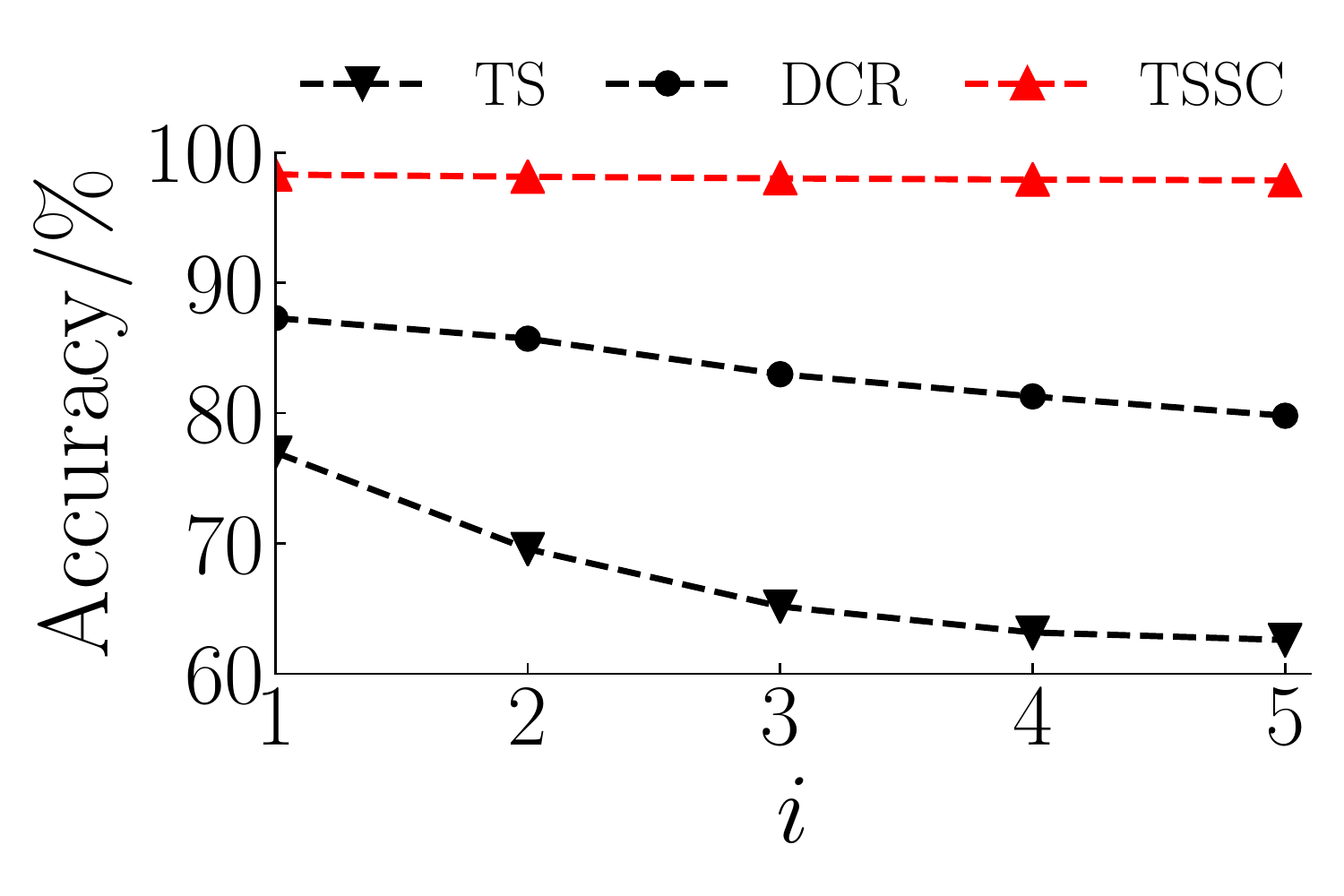}\label{BASE_0}}
\subfloat[]{\includegraphics[width=0.23\textwidth]{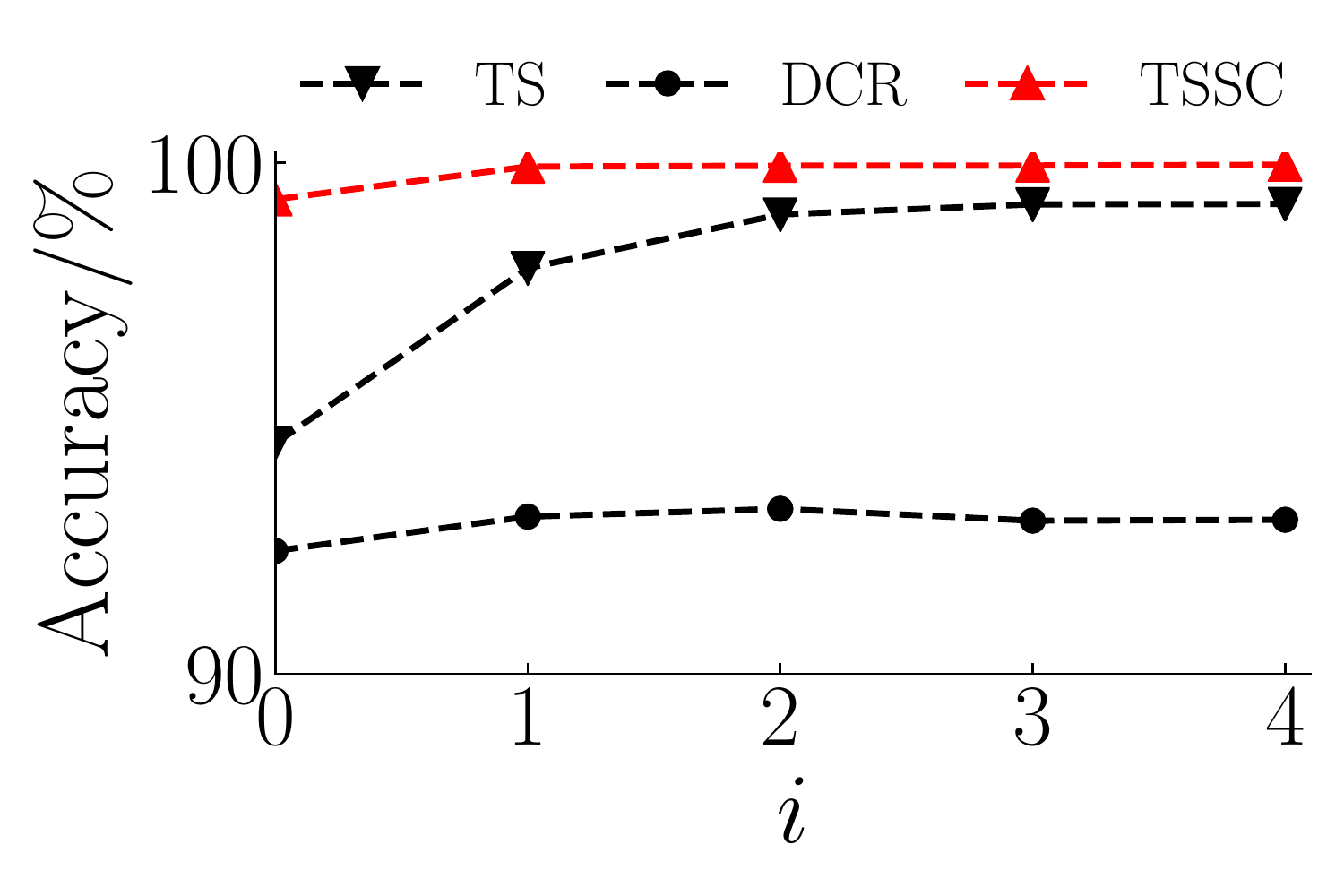}\label{BASE_5}}
\label{fig:butterfly-effect}
\caption{The results of the second experiment (a) the classification accuracies in the first trial with the test datasets, $BASE_i,\; i=1,2,3,4,5$ using ConvNet trained by the dataset $BASE_0$. (b) the classification accuracies in the second trial with the tests with the test datasets, $BASE_i,\; i=0,1,2,3,4$ using ConvNet trained by the dataset $BASE_5$. (All the classification accuracies are listed in Table~\ref{tab:BASE_0} and~\ref{tab:BASE_5} in Supplemental Material)}\label{fig:initial_condition}
\end{figure}

\textit{Results for Initial Conditions}---The sensitive dependence on initial conditions is known as the butterfly effect\cite{butterfly_1963}. To understand the effect of initial conditions, the second experiment has two trials.
The ranges of the uniform distributions for the samples of initial conditions increase with the five datasets, $BASE_i,\; i=1,2,3,4,5$.
The first trial uses the dataset $BASE_0$ for the training of ConvNet and the other five datasets, $BASE_i,\;i=1,2,3,4,5$ for the test. The other trial uses the dataset $BASE_5$ for the training and the other five datasets, $BASE_i,\;i=0,1,2,3,4$ for the test. 
TSSC-ConvNet is very robust and extremely accurate at 100{\%} in both trials.
In Figure~\ref{BASE_0}, the first trial shows that the classification accuracies of TSSC-ConvNet are very high near at 100{\%} and almost the same for the five test datasets $BASE_i,\; i=1,2,3,4,5$. But the classification accuracies of TS-ConvNet and DCR-ConvNet decline with the five tests of $BASE_i,\; i=1,2,3,4,5$ and below 90{\%} and 80{\%} respectively. 
In Figure~\ref{BASE_5}, the second trial shows that ConvNet trained with $BASE_5$ can significantly improve the performance of TS-ConvNet and DCR-ConvNet to the accuracies above 90{\%}. Particularly the classification accuracies of TS-ConvNet is even better than DCR-ConvNet in the second trial while the classification accuracies of TS-ConvNet is worse than DCR-ConvNet below 80{\%} in the first trial.
The performance of TSSC-ConvNet is very robust in both trials since the patterns in the TSSC images are indifferent to the samples of initial conditions. But the two trials show that the classification accuracies of TS-ConvNet and DCR-ConvNet are both very sensitive to the samples of initial conditions in the training dataset, {i.e.}, which training dataset is used to train ConvNet can affect the classification performance for TS-ConvNet and DCR-ConvNet.  

\begin{figure}[H]
\centering
\subfloat[]{\includegraphics[width=0.23\textwidth]{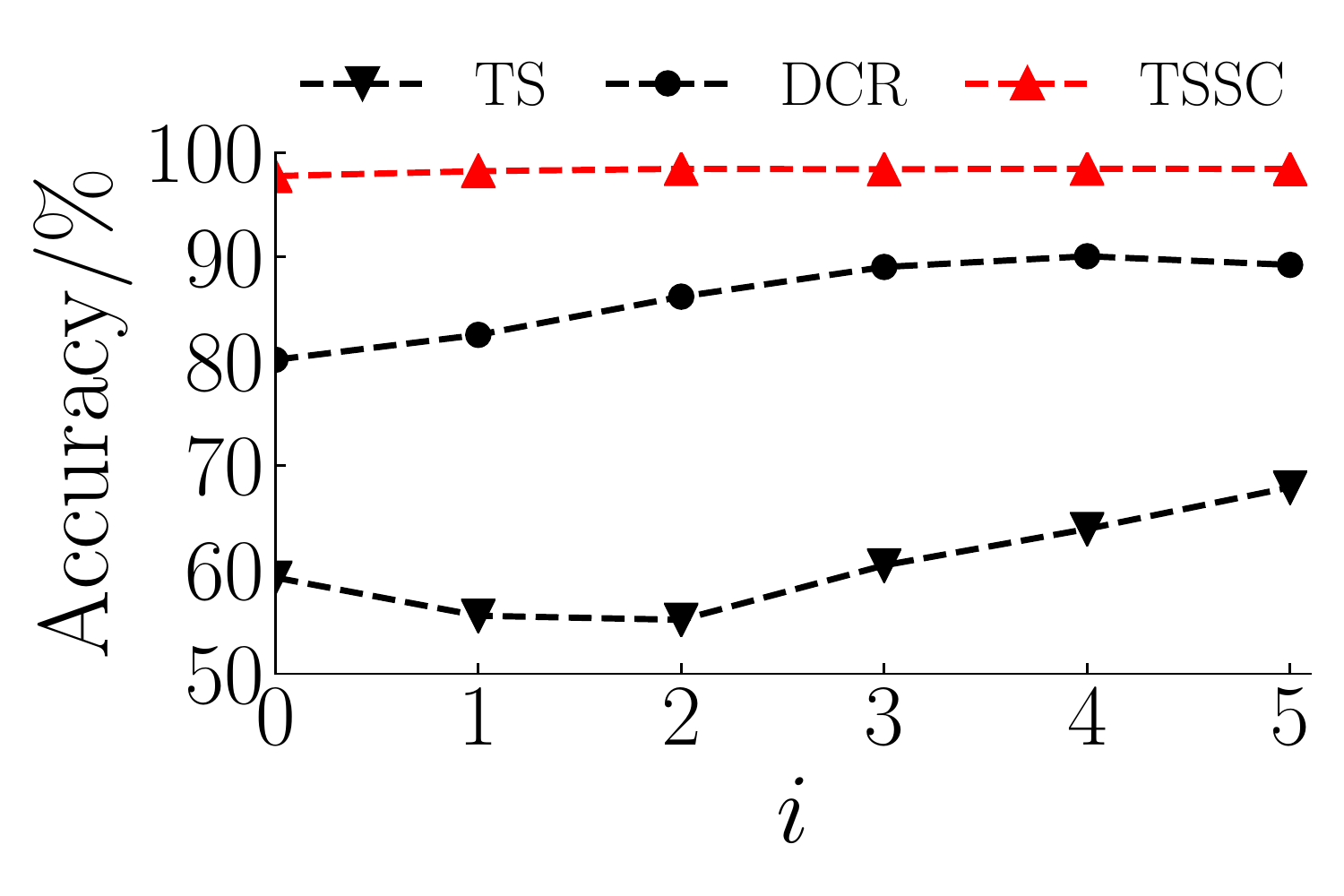}\label{NS_SP}}
\subfloat[]{\includegraphics[width=0.23\textwidth]{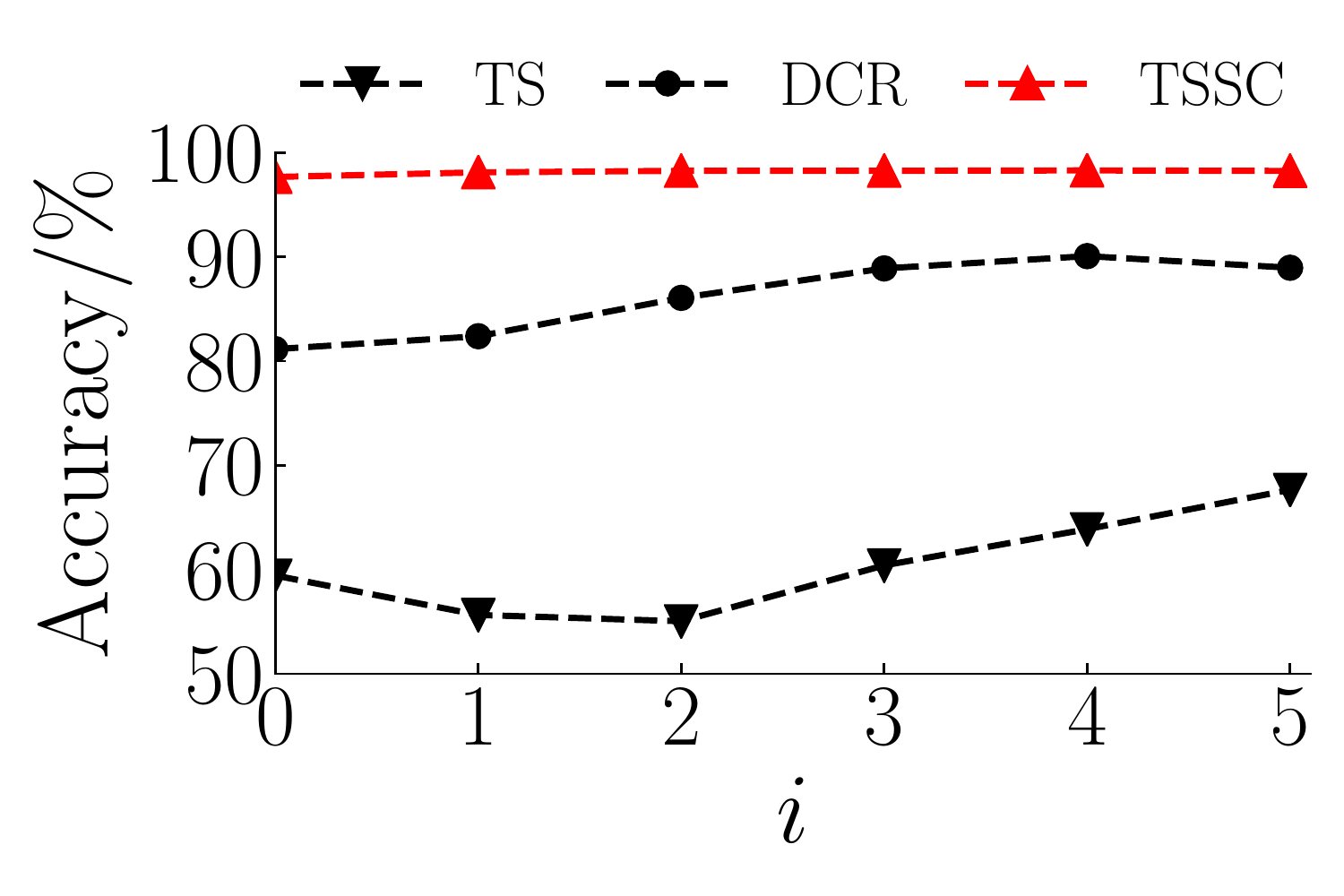}\label{NS_DP}}
\caption{The results of third experiment (a) the classification accuracies in the first trial with the test datasets, $NS^{SP}_i$ using ConvNet trained by the dataset $BASE_i$. (b) the classification accuracies in the second trial with the test datasets, $NS^{DP}_i$ using ConvNet trained by the dataset $BASE_i$. (All the classification accuracies are listed in Table~\ref{tab:NS_SP} and~\ref{tab:NS_DP} in Supplemental Material)}\label{fig:nonstationary}
\end{figure}

\textit{Results for Segmentations}---The segmentation of finite length time series can make the classification complicated given the non-stationary and nonlinear nature of chaotic dynamics. In the third experiment, the six datasets, $BASE_i,\; i=0,1,2,3,4,5$ are used to train ConvNet. With the corresponding datasets $NS^{SP}_i$ and $NS^{DP}_i$, two trials are carried out for the test and benchmark of the three classifiers, TS-ConvNet, DCR-ConvNet and TSSC-ConvNet. $BASE_i$ and $NS^{SP}_i$ are the two segments of the same time series while $NS^{DP}_i$ and $DP_i$ are the two segments of the same time series. Using $NS^{SP}_i$, the first trial is the test of the segmentation effect with the same control parameter samples. Using the $NS^{DP}_i$, the second trial is the test of the segmentation effect with the different control parameter samples.
Figures~\ref{NS_SP} and \ref{NS_DP} have very similar results for all three classifiers for the trials. Particularly, as in the first and second experiments, TSSC-ConvNet still has the best classification accuracies near at 100{\%} and robust performance in the third experiment. Both TS-ConvNet and DCR-ConvNet are very sensitive to the segmentations and have much poorer performance. The classification accuracies of TS-ConvNet are between 50{\%} and 70{\%} and the ones of DCR-ConvNet are between 80{\%} and 90{\%}. In addition, the performances of TS-ConvNet and DCR-ConvNet are sensitive to the samples of initial conditions of the training datasets $BASE_i,\;i=0,1,2,3,4,5$. 

\textit{Conclusion}---The novel TSSC image encoding method of time series reveals new triad clustering patterns. The existing practical time series classification methods don't work well for the classification of chaotic time series. Based on the TSSC images, ConvNet demonstrates superior accuracy and robust performance for the chaotic time series classification in benchmark with the raw time series and the popular image encoding method DCR. More interesting, In the future, the TSSC image can be extended for the unsupervised classification of time series by exploring the image-based similarity measure. Also, the application of higher order time series motifs beyond triads can help us identify more important temporal structures in time series and enhance the ability of classification and prediction.
\begin{acknowledgments}
Xin Chen acknowledges the financial support from the National Natural Science Foundation of China (Grant No. 21773182 (B030103) ). 
\end{acknowledgments}

\pagebreak
\widetext
\begin{center}
\textbf{\large Supplemental Materials: Triad State Space Construction for Chaotic Signal Classification with Deep Learning}
\end{center}
\setcounter{equation}{0}
\setcounter{figure}{0}
\setcounter{table}{0}
\setcounter{page}{1}
\makeatletter
\renewcommand{\theequation}{S\arabic{equation}}
\renewcommand{\thetable}{S\arabic{table}}
\renewcommand{\thefigure}{S\arabic{figure}}
\renewcommand{\bibnumfmt}[1]{[S#1]}
\renewcommand{\citenumfont}[1]{S#1}

\section{ConvNet Structure}\label{CS}
\begin{figure}[H]
\centering
\includegraphics[width=0.4\textwidth]{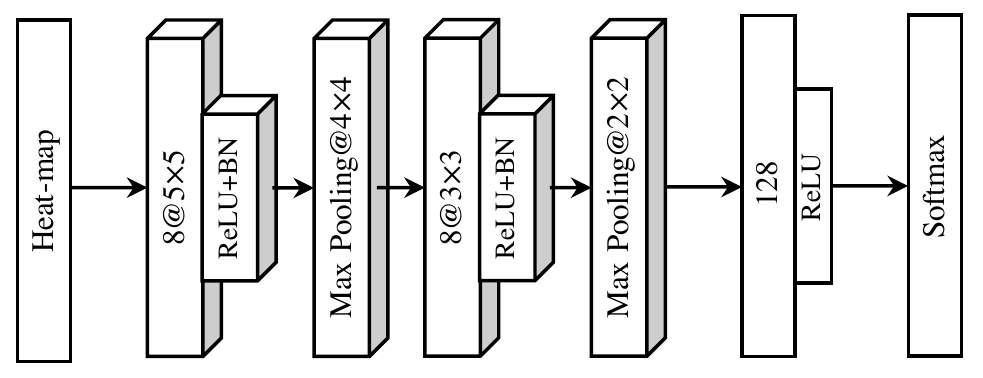}
\caption{The ConvNet diagram with the input of TSSC or DCR heat-map. The formula $8@5\times5$ means the convolutional layer has the filter size of $8$ and the receptive field $5\times5$. BN is Batch Normalization.}\label{fig:heatmap-convnet}
\end{figure}

\begin{figure}[H]
\centering
\includegraphics[width=0.4\textwidth]{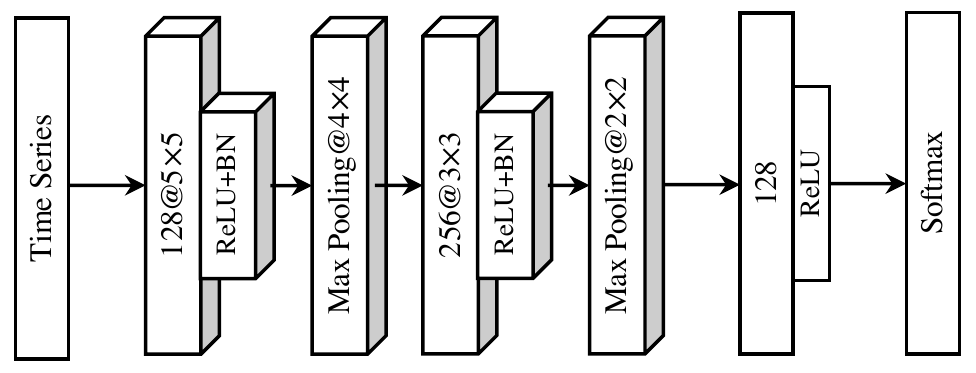}
\caption{The ConvNet diagram with the input of raw time series. The formula $128@5\times5$ means the convolutional layer has the filter size of $128$ and the receptive field $5\times5$. BN is Batch Normalization.}\label{fig:ts-convnet}
\end{figure}

Figure~\ref{fig:heatmap-convnet} shows the structure of DCR-ConvNet and TSSC-ConvNet.
Figure~\ref{fig:ts-convnet} shows the structure of TS-ConvNet. 

\section{TSSC and DCR Images}\label{TDI}
\begin{figure}[H]
\centering
\subfloat[Logistic map ($r=4.0$)]{\includegraphics[width=0.23\textwidth]{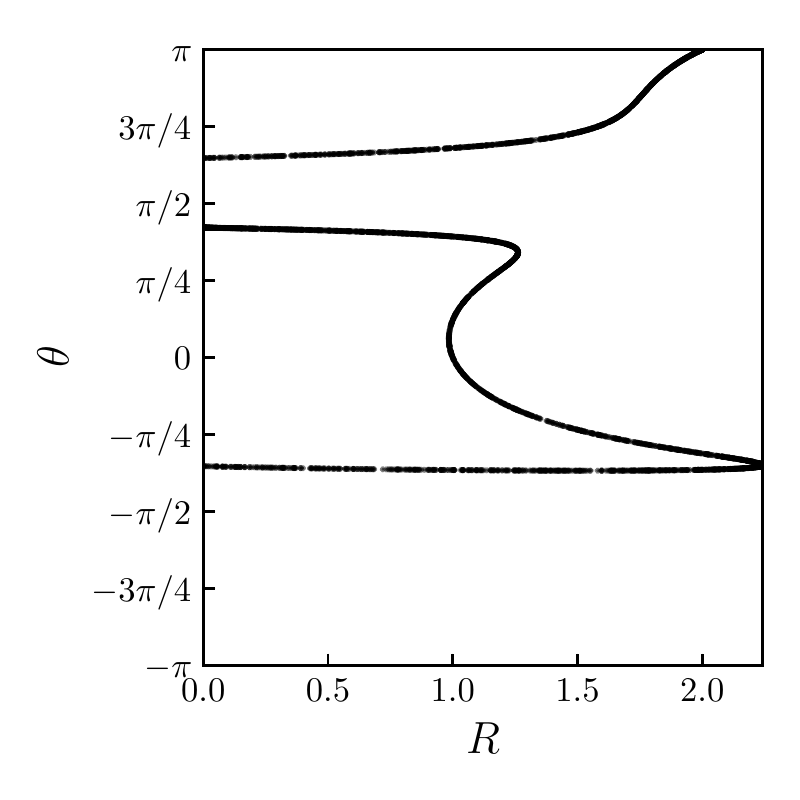}} 
\subfloat[Linear congruential generator ($C=259200$)]{\includegraphics[width=0.23\textwidth]{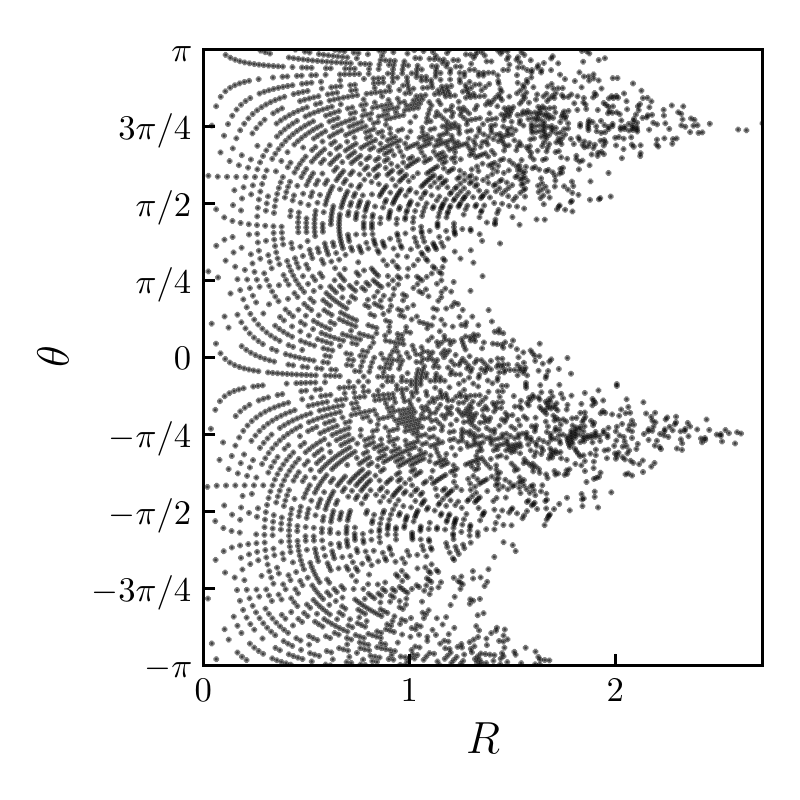}} 
\subfloat[Skew tent map  ($w=0.8$)]{\includegraphics[width=0.23\textwidth]{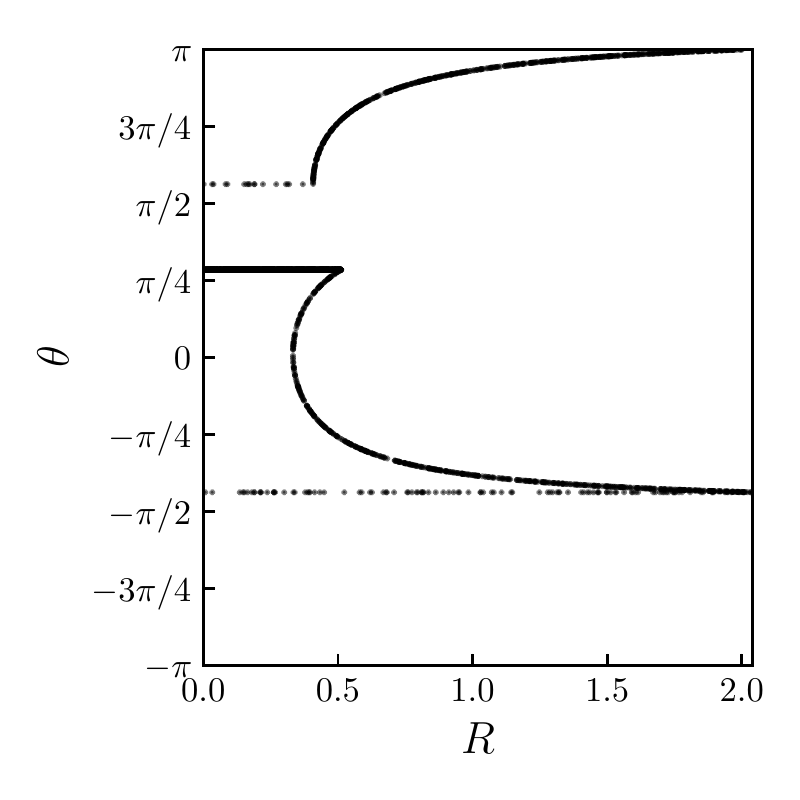}} \\
\subfloat[Lozi map ($a=1.7, b=0.5$)]{\includegraphics[width=0.23\textwidth]{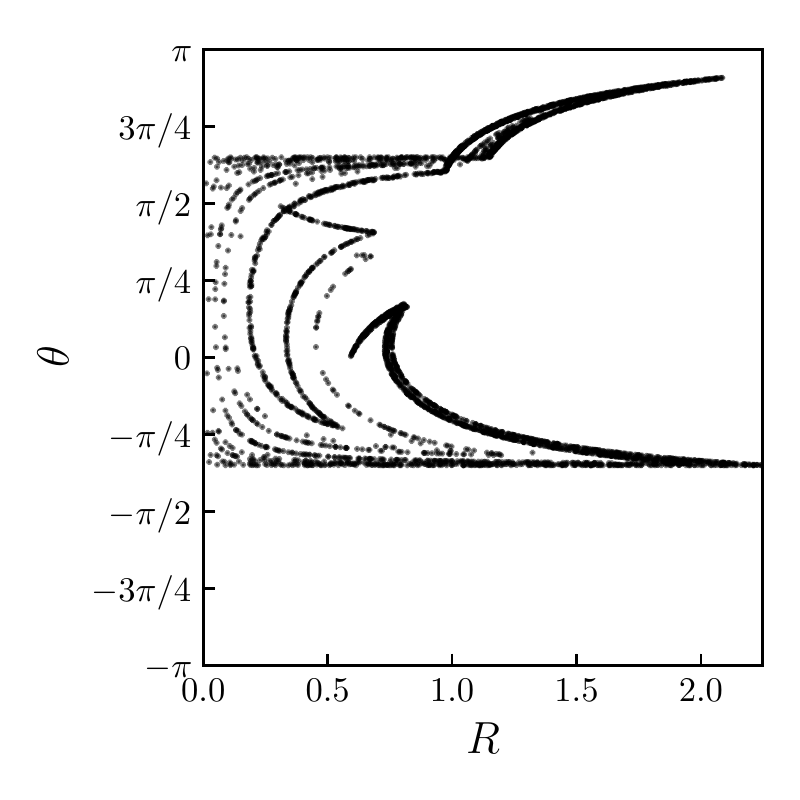}} 
\subfloat[Dissipative standard map ($b=0.1, k=8.8$)]{\includegraphics[width=0.23\textwidth]{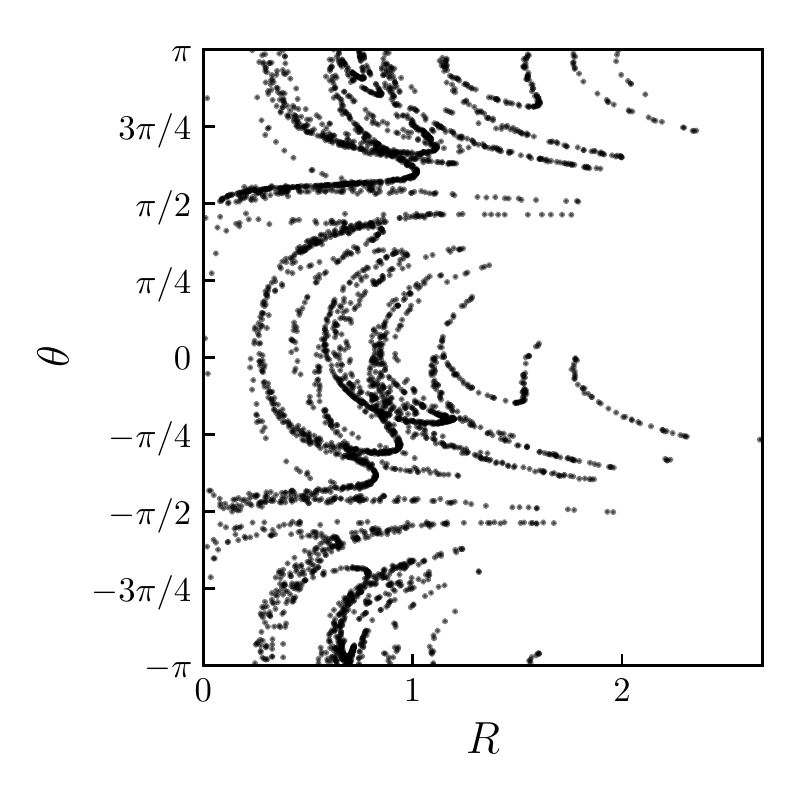}}
\subfloat[Sinai map ($\delta=0.1$)]{\includegraphics[width=0.23\textwidth]{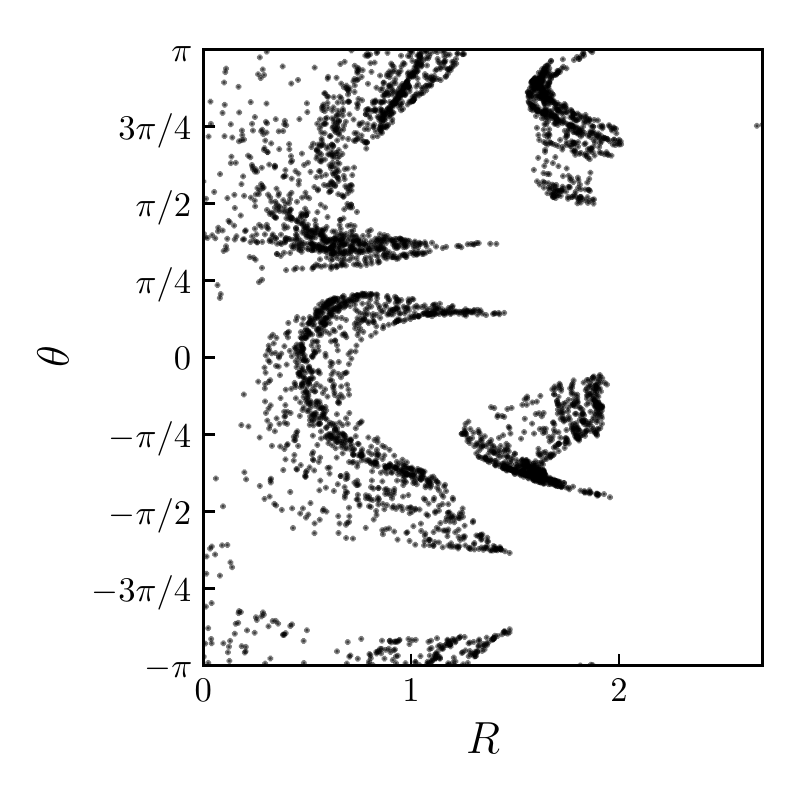}}\\
\subfloat[Annode's cat map ($k=2$)]{\includegraphics[width=0.23\textwidth]{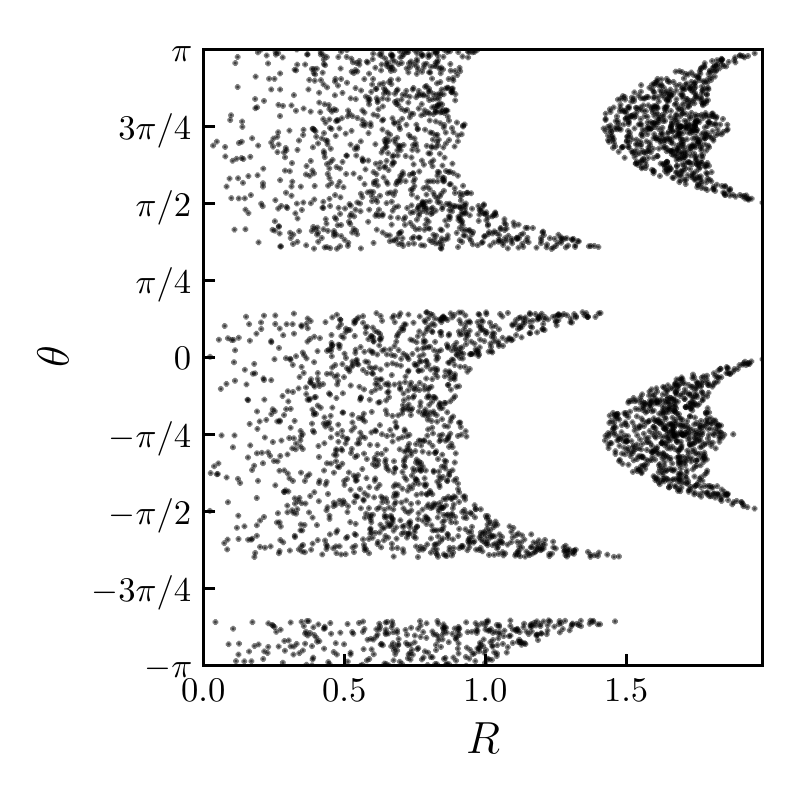}}
\subfloat[Chirikov standard map ($k=1.0$)]{\includegraphics[width=0.23\textwidth]{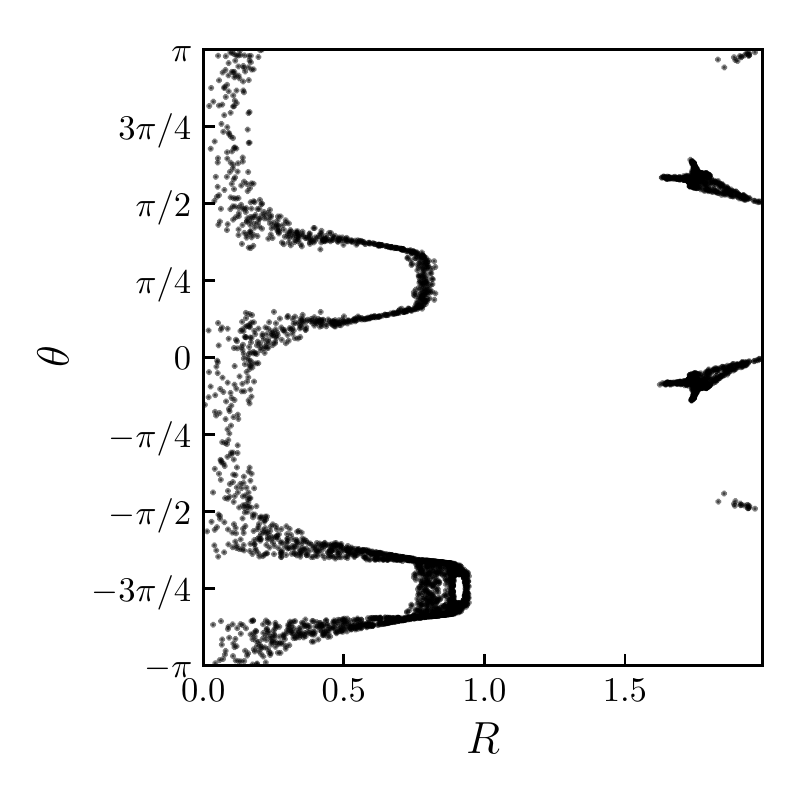}} 
\subfloat[Chaotic web map ($k=1.0$)]{\includegraphics[width=0.23\textwidth]{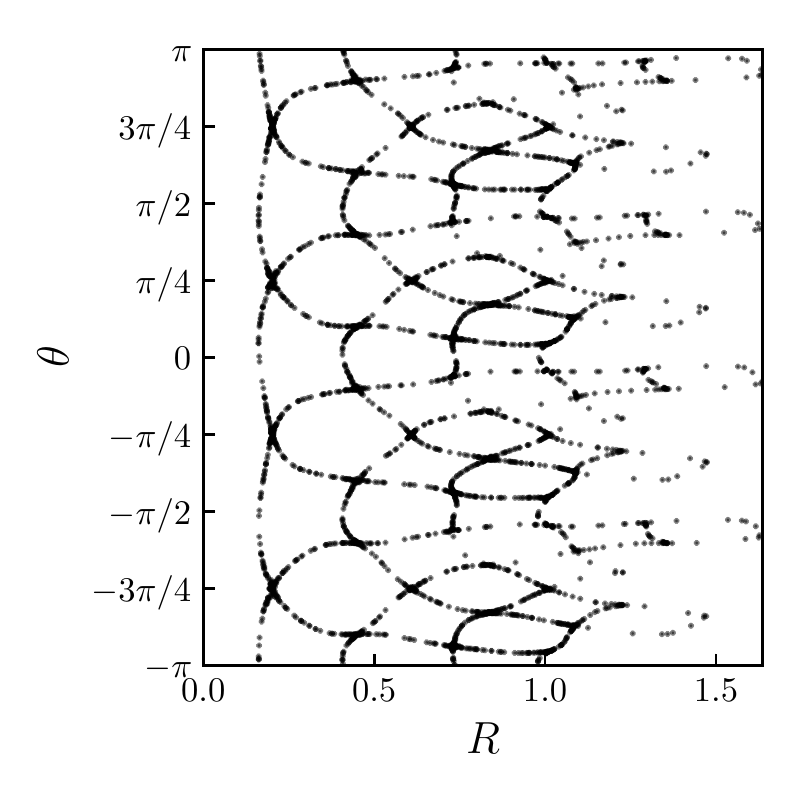}}\\
\caption{TSSC images of the nine chaotic dynamics maps. The initial conditions of the nine chaotic maps are listed in the Table~\ref{tab:TSsetting}. Each time series has $4000$ steps normalized to the range of $[-1, 1]$.}
\label{fig:chaotic-TSS}
\end{figure}

\begin{figure}[H]
\centering
\subfloat[Logistic map ($r=4.0$)]{\includegraphics[width=0.23\textwidth]{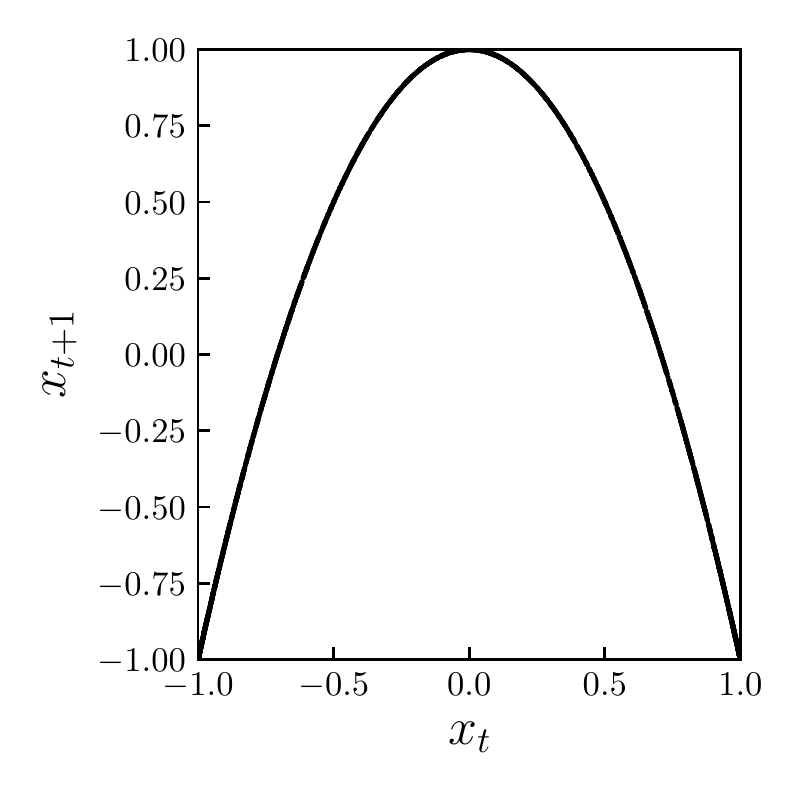}} 
\subfloat[Linear congruential generator ($C=259200$)]{\includegraphics[width=0.23\textwidth]{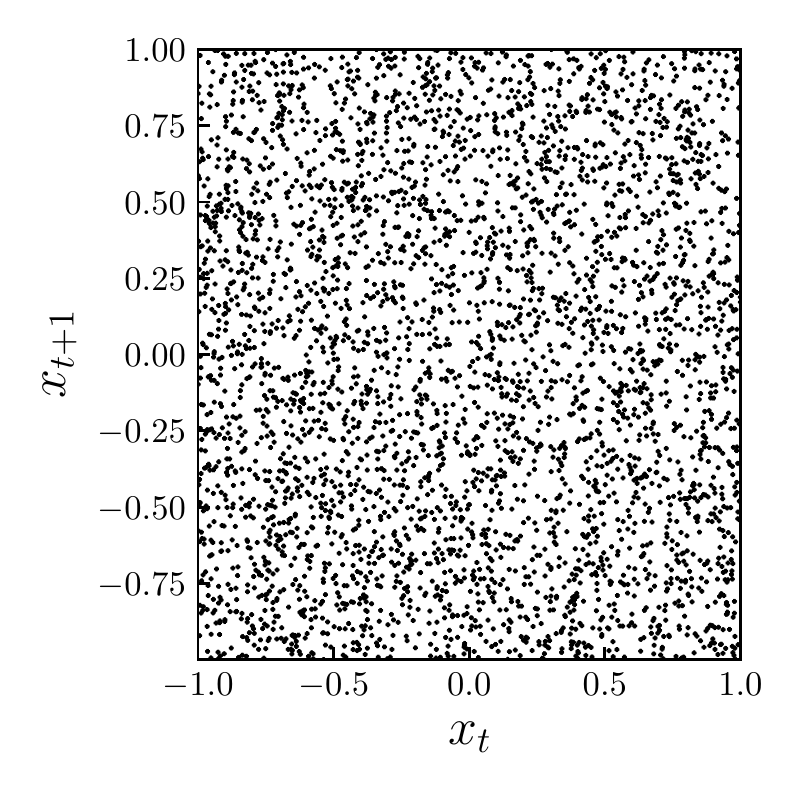}}
\subfloat[Skew tent map  ($w=0.8$)]{\includegraphics[width=0.23\textwidth]{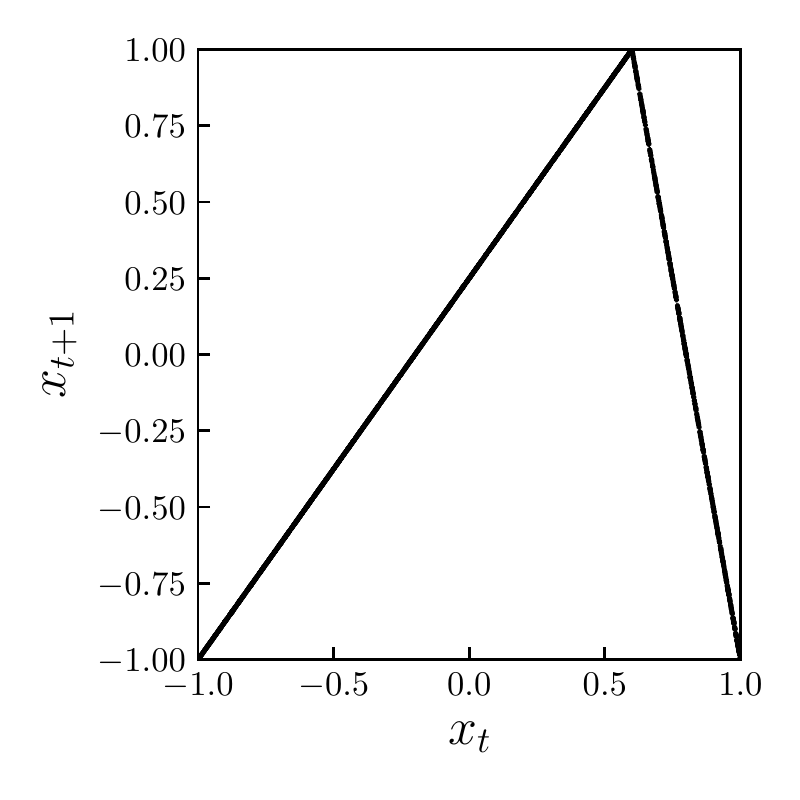}} \\
\subfloat[Lozi map ($a=1.7, b=0.5$)]{\includegraphics[width=0.23\textwidth]{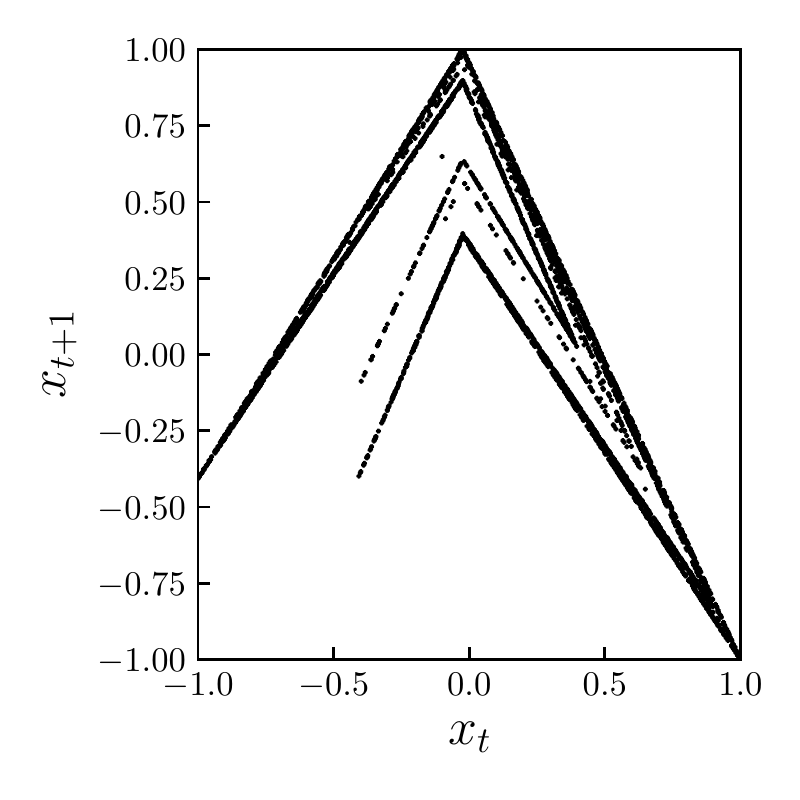}} 
\subfloat[Dissipative standard map ($b=0.1, k=8.8$)]{\includegraphics[width=0.23\textwidth]{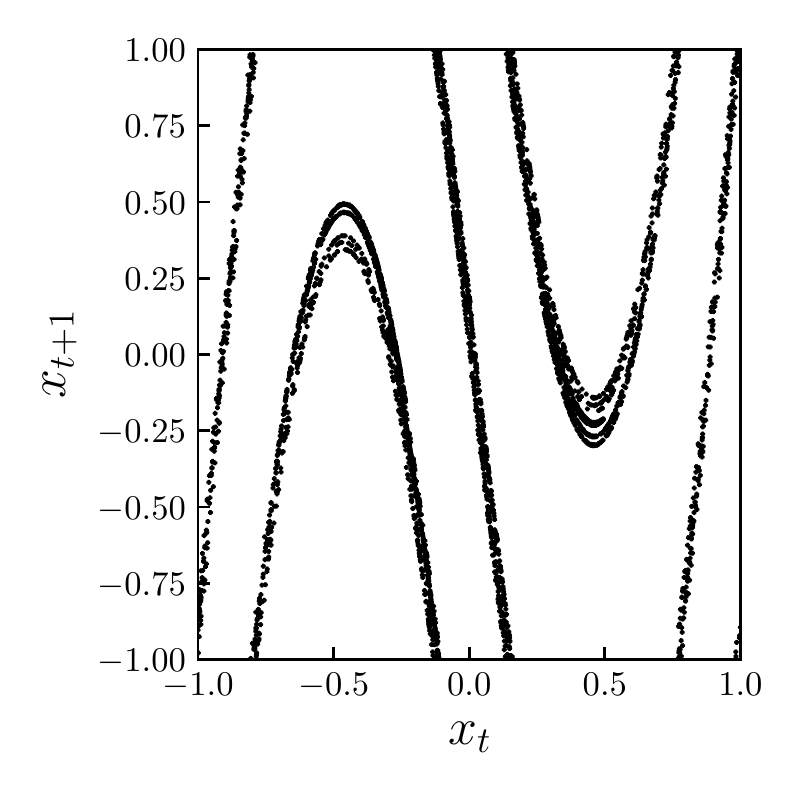}}
\subfloat[Sinai map ($\delta=0.1$)]{\includegraphics[width=0.23\textwidth]{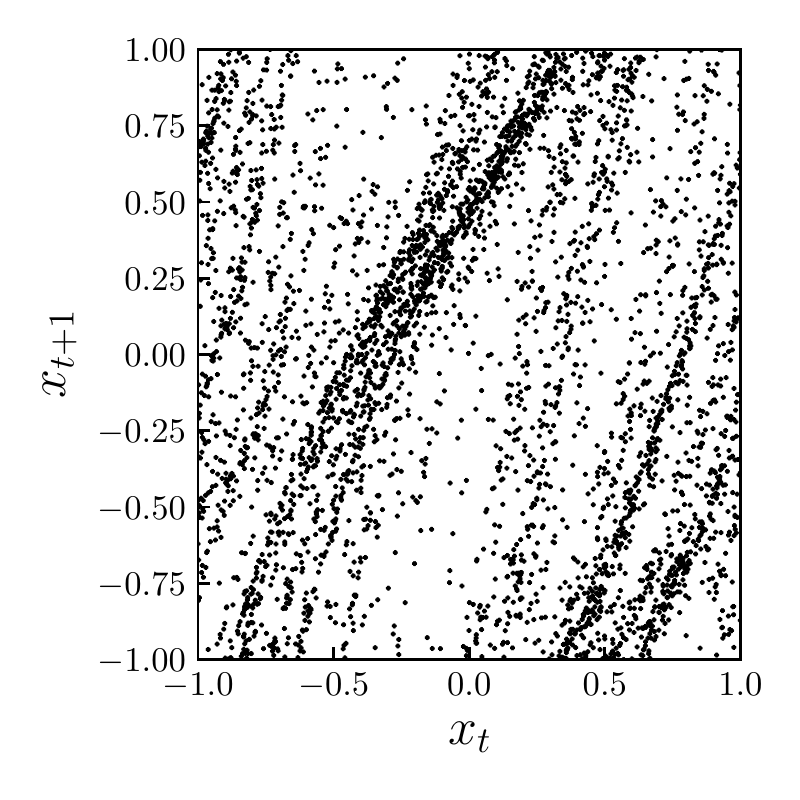}}\\
\subfloat[Annode's cat map ($k=2$)]{\includegraphics[width=0.23\textwidth]{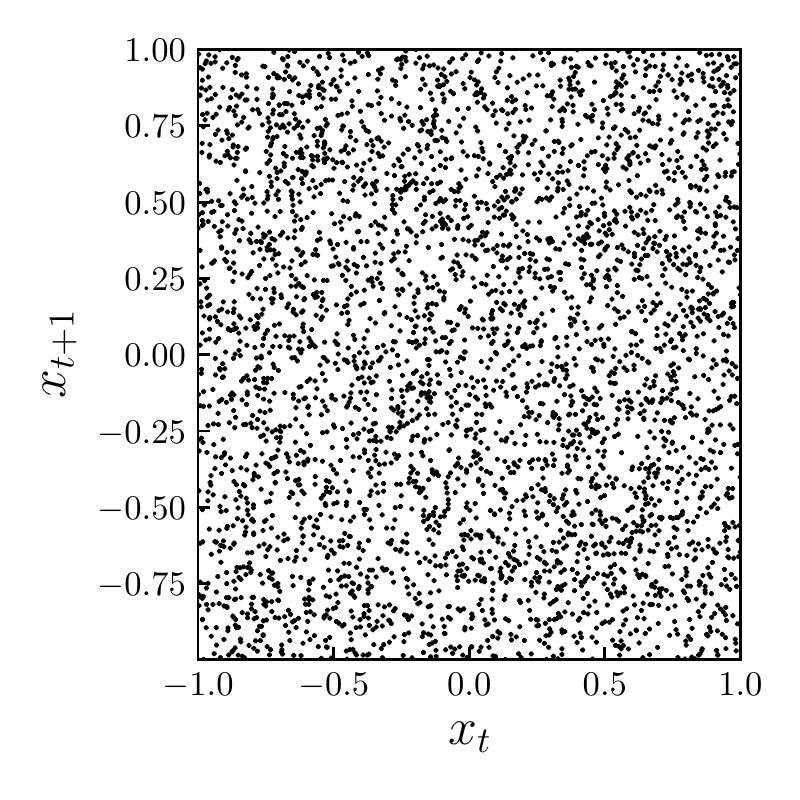}}
\subfloat[Chirikov standard map ($k=1.0$)]{\includegraphics[width=0.23\textwidth]{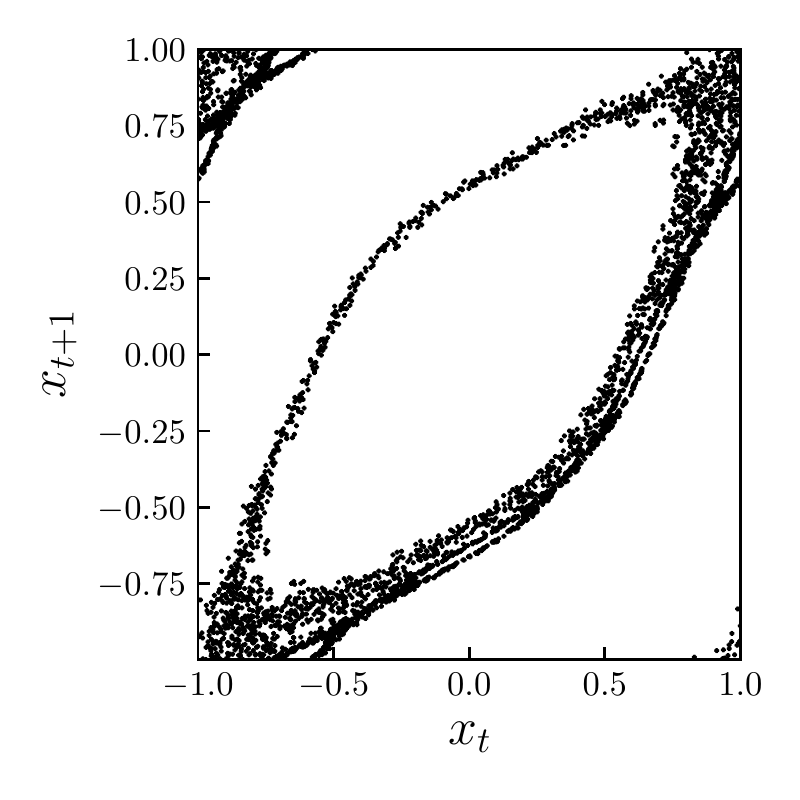}} 
\subfloat[Chaotic web map ($k=1.0$)]{\includegraphics[width=0.23\textwidth]{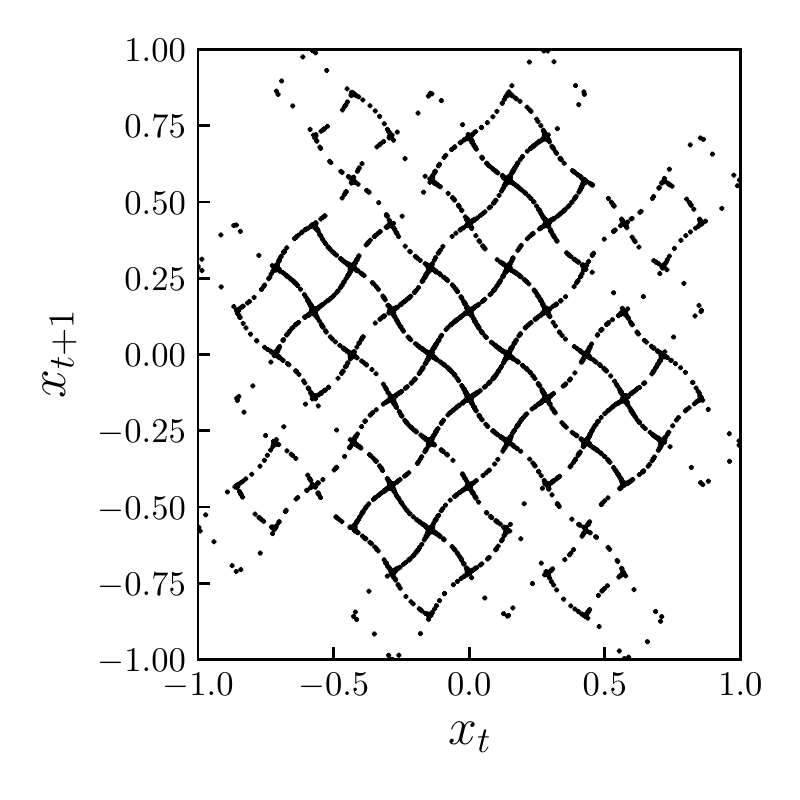}}
\\
\caption{DCR images of the nine chaotic maps whose formulas and initial conditions are in Table~\ref{tab:TSsetting}. Each time series has $4000$ steps normalized to $[-1, 1]$.}
\label{fig:chaotic-DCR}
\end{figure}

Figures~\ref{fig:chaotic-TSS} and~\ref{fig:chaotic-DCR} shows the TSSC and DCR images of the surrogate time series of nine chaotic dynamic maps. 

\section{Control Parameters and Initiail Conditions}
The control parameters and initial conditions of the nine chaotic maps for the dataset $D_0$ in Figure~\ref{fig:dataset} are presented in Table~\ref{tab:TSsetting}.
\begin{table}[H]
 \caption{Control parameters and initials conditions of the nine chaotic maps. ($N, D, C$ are short for the non-invertible, dissipative, conservative maps.)}
  \centering
  \begin{ruledtabular}
  \begin{tabular}{llll}
     \tabincell{l}{Name} & Evolution Function & \tabincell{l}{Control\\ Parameters (CP)} & Initial Conditions (IC) \\
    \colrule
    Logistic map$^N$  & $x_{t+1}=rx_t(1-x_t)$   & $r\in[3.5, 4.0)$  & $x_1=10^{-6}$ \\
    \tabincell{l}{Linear congruential\\ generator$^N$ }   & $x_{t+1}=Ax_t+B \pmod{C}$ & \tabincell{l}{$A=7141$\\$B=54773$\\$C\in[259200.0, $\\$ 600000.0)$}  &   $x_1=0.0$ \\
    Skew tent map$^N$  & $x_{t+1}=\left\{
        \begin{array}{lcl}
        x_t/w       &      & {x<=w}\\
        (1-x_t)/(1-w)     &      & {x>w}
        \end{array} \right.$ &  $w\in[0.11, 0.9)$ & $x_1=0.1$    \\
    Lozi map$^D$  & \tabincell{l}{$x_{t+1}=1-a |x_t| +b y_t$\\$y_{t+1}=x_t$} &  \tabincell{l}{$a\in(1.6, 1.8)$ \\ $b\in(0.4, 0.6)$} & \tabincell{l}{$x_1=-0.1$ \\ $y_1=0.1$} \\
    \tabincell{l}{Dissipative\\ standard map$^D$  }  & \tabincell{l}{$x_{t+1}=x_t+y_{t+1} \pmod{2\pi}$\\$y_{t+1}=by_t+k \sin x_t \pmod{2\pi}$} &  \tabincell{l}{$b\in(0.1, 1.0)$ \\ $k\in(1.0, 10.0)$} & \tabincell{l}{$x_1=0.1$ \\ $y_1=0.1$} \\
    Sinai map$^D$  & \tabincell{l}{$x_{t+1}=x_t+y_t+\delta \cos 2\pi y_t \pmod{1}$\\$y_{t+1}=x_t+2y_t \pmod{1}$} &  $\delta\in(0.1, 1.0)$ & \tabincell{l}{$x_1=0.9$ \\ $y_1=0.5$}  \\
    Annode's cat map$^C$   & \tabincell{l}{$x_{t+1}= x_t+y_t \pmod{1}$\\$y_{t+1}=x_t+k y_t \pmod{1}$} &  $k\in[1.0, 10.0)$ & \tabincell{l}{$x_1=0.0$ \\ $y_1 = 1/\sqrt{2}$} \\
    Chirikov standard map$^C$  & \tabincell{l}{$x_{t+1}=x_t+y_{t+1} \pmod{2\pi}$\\$y_{t+1}=y_t+k \sin x_t \pmod{2\pi}$} &  $k\in[1.0, 5.0)$ & \tabincell{l}{$x_1=0.0$ \\ $y_1 = 6.0$} \\
    Chaotic web map$^C$ & \tabincell{l}{$x_{t+1}=x_t \cos \alpha - (y_t+k\sin x_t) \sin \alpha$\\$y_{t+1}=x_t \sin \alpha + (y_t+k\sin x_t) \cos \alpha$} &  $k\in[1.0, 5.0)$ & \tabincell{l}{$x_1=0.0$ \\ $y_1 = 3.0$}
  \end{tabular}
  \end{ruledtabular}
  \label{tab:TSsetting}
\end{table}

\section{Benchmark Results on Control Parameters, Initial Conditions and Segmentations}
Tables~\ref{tab:DP} shows the classification accuracies of the first experiment on the control parameter.
\begin{table}[H]
 \caption{With ConvNet trained with the dataset $BASE_i, i=0,1,2,3,4,5$, the classification accuracies according to $DP_i\; i=0,1,2,3,4,5$ respectively.} \centering
  \begin{ruledtabular}
  \begin{tabular}{cccc}
    $i$  & TS-ConvNet   & DCR-ConvNet   &TSSC-ConvNet                \\
    \colrule
    0 &   92.7 &   91.7 &   \textbf{99.3} \\
    1 &   99.5 &   92.9 &   \textbf{99.8} \\
    2 &   99.6 &   93.0 &   \textbf{99.8} \\
    3 &   99.5 &   93.1 &   \textbf{99.7} \\
    4 &   99.3 &   93.2 &   \textbf{99.7} \\
    5 &   99.0 &   93.0 &   \textbf{99.7} \\
  \end{tabular}
  \end{ruledtabular}
  \label{tab:DP}
\end{table}

Tables~\ref{tab:BASE_0} and~\ref{tab:BASE_5} show the classification accuracies of the two trials in the second experiment on the random samples of initial conditions.
\begin{table}[H]
 \caption{With ConvNet trained with the datasets $BASE_0$, the classification accuracies according to $BASE_i\; i=1,2,3,4,5$.} \centering
  \begin{ruledtabular}
  \begin{tabular}{cccc}
  \toprule
  $i$  & TS-ConvNet   & DCR-ConvNet   &TSSC-ConvNet                \\
  \colrule
  1 &   77.0 &   87.3 &   \textbf{98.3} \\
  2 &   69.6 &   85.7 &   \textbf{98.1} \\
  3 &   65.2 &   83.0 &   \textbf{98.0} \\
  4 &   63.2 &   81.3 &   \textbf{97.9} \\
  5 &   62.6 &   79.8 &   \textbf{97.9} \\
  \bottomrule
  \end{tabular}
  \end{ruledtabular}
  \label{tab:BASE_0}
\end{table}
\begin{table}[H]
 \caption{With ConvNet trained with the dataset $BASE_5$, the classification accuracies based on $BASE_i\;i=0,1,2,3,4$.} \centering
  \begin{ruledtabular}
  \begin{tabular}{cccc}
  \toprule
  $i$  & TS-ConvNet   & DCR-ConvNet   &TSSC-ConvNet                \\
  \colrule
  0 &   94.5 &   92.4 &   \textbf{99.3} \\
  1 &   97.9 &   93.1 &   \textbf{99.9} \\
  2 &   99.0 &   93.2 &   \textbf{99.9} \\
  3 &   99.2 &   93.0 &   \textbf{99.9} \\
  4 &   99.2 &   93.0 &   \textbf{100.0} \\
  \bottomrule
  \end{tabular}
  \end{ruledtabular}
  \label{tab:BASE_5}
\end{table}

Tables~\ref{tab:NS_SP} and~\ref{tab:NS_DP} shows the classification accuracies of the two trials of third experiment on the segmentation. 
\begin{table}[H]
 \caption{According to the test datasets, $NS^{SP}_i,\; i =0,1,2,3,4,5$, the classification accuracies of ConvNet trained with the datasets $BASE_i,\;i=0,1,2,3,4,5$ respectively.} \centering
  \begin{ruledtabular}
  \begin{tabular}{cccc}
  \toprule
  $i$  & TS-ConvNet   & DCR-ConvNet   &TSSC-ConvNet                \\
  \colrule
  0 &   59.2 &   80.1 &   \textbf{97.7} \\
  1 &   55.6 &   82.5 &   \textbf{98.2} \\
  2 &   55.2 &   86.2 &   \textbf{98.4} \\
  3 &   60.4 &   89.0 &   \textbf{98.4} \\
  4 &   63.9 &   90.0 &   \textbf{98.4} \\
  5 &   67.9 &   89.2 &   \textbf{98.4} \\
  \bottomrule
  \end{tabular}
  \end{ruledtabular}
  \label{tab:NS_SP}
\end{table}
\begin{table}[H]
 \caption{According to the test datasets, $NS^{DP}_i,\; i =0,1,2,3,4,5$, the classification accuracies of ConvNet trained with the datasets $BASE_i,\;i=0,1,2,3,4,5$ respectively.} \centering
  \begin{ruledtabular}
  \begin{tabular}{cccc}
  \toprule
  $i$  & TS-ConvNet   & DCR-ConvNet   &TSSC-ConvNet \\
  \colrule
  0 &   59.4 &   81.1 &   \textbf{97.7} \\
  1 &   55.6 &   82.4 &   \textbf{98.1} \\
  2 &   55.0 &   86.1 &   \textbf{98.2} \\
  3 &   60.4 &   88.9 &   \textbf{98.2} \\
  4 &   63.9 &   90.1 &   \textbf{98.3} \\
  5 &   67.6 &   88.9 &   \textbf{98.2} \\
  \bottomrule
  \end{tabular}
  \end{ruledtabular}
  \label{tab:NS_DP}
\end{table}

\end{document}